\documentclass[iicol]{sn-jnl}

\usepackage{graphics} 
\usepackage{epsfig} 
\usepackage{mathptmx} 
\usepackage{times} 
\usepackage{amsmath} 
\usepackage{amssymb}  
\usepackage{algorithm}
\usepackage{algpseudocode}
\usepackage{xcolor}
\usepackage{multirow}
\usepackage[latin1]{inputenc}

\usepackage{etoolbox}
    \makeatletter
    \patchcmd{\ps@headings}
    {\hbox to \hsize{\hfill Springer Nature 2021 \LaTeX\ template\hfill}}
    {\hbox to \hsize{\hfill Based on  Springer Nature \LaTeX\ template\hfill}}
    {}
    {}
    \patchcmd{\ps@titlepage}
    {\hbox to \hsize{\hfill Springer Nature 2021 \LaTeX\ template\hfill}}
    {\hbox to \hsize{\hfill Based on  Springer Nature \LaTeX\ template\hfill}}
    {}
    {}
    \makeatother

\newcommand{\al}{\alpha}

\newcommand{\xbj}[2]{X_{#1}^{#2}}
\newcommand{\hxk}{\hat{X}_k}        
\newcommand{\xd}[1]{X_{#1}^{i^{\dagger}}}        
\newcommand{\qk}{q_k}               
\newcommand{\gbkmbk}{g_{k-1, k}}    
\newcommand{\gbobi}{g_{0, i}}       
\newcommand{\gbibj}{g_{i, j}}       
\newcommand{\gbobone}{g_{0, 1}}       
\newcommand{\dotgbobi}{\dot{g}_{0, i}}
\newcommand{\hVbobi}{\hat{V}_{0, i}^{i}}
\newcommand{\Vbobi}{V_{0, i}^{i}}
\newcommand{\Jbobi}{J_{0, i}^{i}}
\newcommand{\Rbobi}{R_{0, i}}
\newcommand{\dotRbobi}{\dot{R}_{0, i}}
\newcommand{\homegabobi}{\hat{\omega}_{0, i}^{i}}
\newcommand{\vbobi}{v_{0, i}^{i}}
\newcommand{\dotdbobi}{\dot{d}_{0, i}^{i}}
\newcommand{\wbj}[2]{W_{#1}^{#2}}
\newcommand{\Sf}{\{S\}}
\newcommand{\Gf}{\{G\}}
\newcommand{\Hf}{\{H\}}
\newcommand{\Vgs}{{V}_{G, S}^{S}}
\newcommand{\dsda}{\norm{\bp_{,\al}}}
\newcommand{\Voone}{{V}_{0, 1}^{1}}
\newcommand{\Joone}{{J}_{0, 1}^{1}}
\newcommand{\dsdqone}{\norm{\bp_{,q_1}}}
\newcommand{\Bo}{\{B_{0}\}}
\newcommand{\Bone}{\{B_{1}\}}
\newcommand{\Btwo}{\{B_{2}\}}
\newcommand{\Bthree}{\{B_{3}\}}

\newcommand{\Bsix}{\{B_{6}\}}
\newcommand{\Bk}{\{B_{k}\}}
\newcommand{\Bkmone}{\{B_{k-1}\}}
\newcommand{\Bi}{\{B_{i}\}}
\newcommand{\Bj}{\{B_{j}\}}

\newcommand{\WS}{W_{k_{\text{J}}}^{k}}
\newcommand{\WE}{W_{k_{\text{E}}}^{k}}

\newcommand{\sadvi}{\ad_{V_{k}^{k}}^{*}}
\newcommand{\der}{\text{d}}

\newcommand{\glambda}{g_{k-1, k}}
\newcommand{\Vlambda}{V_{\lambda}^{\lambda}}
\newcommand{\mAdlambda}{\Ad_{\glambda}^{-1}}
\newcommand{\Vk}{{V_{k}}^{k}}
\newcommand{\hMk}{\hat{M}_{k}^{k}}
\newcommand{\hbk}{\hat{b}_{k}^{k}}
\newcommand{\Mk}{M_{k}^{k}}
\newcommand{\bk}{b_{k}^{k}}

\newcommand{\bMl}{\bar{M}_{l}^{l}}
\newcommand{\hMl}{\hat{M}_{l}^{l}}
\newcommand{\Alambda}{\Ad_{g_{k+1,k}}}
\newcommand{\WEk}{W_{k_{\text{E}}}^{k}}
\newcommand{\hbl}{\hat{b}_{l}^{l}}
\newcommand{\adl}{\ad_{X_{l}\dot{q}_{l}}}
\newcommand{\adk}{\ad_{X_{k}\dot{q}_{k}}}
\newcommand{\Vkl}{V_k^{l}}
\newcommand{\bbl}{\bar{b}_{l}^{l}}
\newcommand{\WEaero}{W_{6_{\text{E}}}^{6}}
\newcommand{\sAdts}{\Ad_{g_{6,3}}^{*}}
\newcommand{\WEaerot}{W_{6_{\text{E}}}^{3}}
\newcommand{\dVk}{\dot{V_{k}}^{k}}
\newcommand{\dVlk}{\dot{V}_{\lambda}^{\lambda}}

\newcommand{\bm}{\mathbf{m}}
\newcommand{\bn}{\mathbf{n}}      

\newcommand{\bp}{\mathbf{p}}

\newcommand{\bt}{\mathbf{t}}

\newcommand{\bv}{\mathbf{v}}
\newcommand{\bw}{\mathbf{w}}

\DeclareMathOperator{\Ad}{Ad}
\DeclareMathOperator{\ad}{ad}

\renewcommand{\div}{\operatorname{div}} 

\providecommand{\norm}[1]{\lVert#1\rVert}  

\renewcommand{\p}{^{\!\cdot\!}}  

\jyear{2021}%

\theoremstyle{thmstyleone}%
\theoremstyle{thmstyletwo}%

\theoremstyle{thmstylethree}%

\begin{document}

\title[Article Title]{A Lie Group-Based Race Car Model for Systematic Trajectory Optimization on 3D Tracks}


\author*[1]{\fnm{Lorenzo} \sur{Bartali}}\email{lorenzo.bartali@phd.unipi.it}

\author[1]{\fnm{Marco} \sur{Gabiccini}}\email{marco.gabiccini@unipi.it}

\author[1]{\fnm{Eugeniu} \sur{Grabovic}}\email{eugeniu.grabovic@phd.unipi.it}

\author[1]{\fnm{Massimo} \sur{Guiggiani}}\email{massimo.guiggiani@unipi.it}

\affil*[1]{\orgdiv{Dipartimento di Ingegneria Civile e Industriale}, \orgname{Universit\`{a} di Pisa}, \orgaddress{\street{Largo Lucio Lazzarino 1}, \city{Pisa}, \postcode{56122}, \country{Italy}}}

%


\abstract{
In this paper we derive the dynamic equations of a race-car model via Lie-group methods. Lie-group methods are nowadays quite familiar to computational dynamicists and roboticists, but their diffusion within the vehicle dynamics community is still limited. We try to bridge this gap by showing that this framework merges gracefully with the Articulated Body Algorithm (ABA) and enables a fresh and systematic formulation of the vehicle dynamics. A significant contribution is represented by a rigorous reconciliation of the ABA steps with the salient features of vehicle dynamics, such as road-tire interactions, aerodynamic forces and load transfers.



 The proposed approach lends itself both to the definition of direct simulation models and to the systematic assembly of vehicle dynamics equations required, in the form  of equality constraints, in numerical optimal control problems. We put our approach on a test in the latter context which involves the solution of minimum lap-time problem (MLTP). 
 More specifically, a MLTP for a race car on the N\"{u}rburgring circuit is systematically set up with our approach. The equations are then discretized with the \emph{direct collocation} method and solved within the CasADi optimization suite. Both the quality of the solution and the computational efficiency demonstrate the validity of the presented approach.}




\maketitle

\section{Introduction}\label{sec:intro}

Minimum lap time problems are among the hottest topics in the automotive research field. In fact, tools for their solution are nowadays widely employed by automotive industries to investigate car performances and provide guidelines both in the design and tuning stages.

Two fundamental elements of MLTPs are car and track models. Track model choices are closely related to the MLTP formulation, which can be defined in a time or spatial domain. As well described in~\cite{Massaro-general}, the latter approach is the most commonly used even if it requires a well defined and differentiable track. Hence, a spline representation is often used and the state of the art is well represented by~\cite{Track-Ribbon} and~\cite{Track-Ribbon-camber}. In~\cite{Track-Ribbon} a 3D ribbon shaped race track model is obtained using a generalized Frenet-Serret apparatus. In particular, the authors propose an optimal estimation procedure that provides a smooth parametrization of the road from noisy data, allowing to model curvature, camber and elevation changes, as well as a variable track width.
Instead, Lovato et al. in~\cite{Track-Ribbon-camber} extended the laterally-flat ribbon-type road model to include lateral curvature. This accounts for lateral camber variations across the track.
Hence, lateral position-dependent camber is introduced as a generalisation required for some race tracks.

The choice of the car model depends on the level of details required to describe the vehicle dynamics. The most simplified model is the single-track one~\cite{Guiggiani}.
Rucco et al.~\cite{SingleTrack} formulate an optimal control problem adopting the single-track model on a 2D track, and include important aspects of vehicle dynamics such as load transfers and nonlinear tire models.
Increasing in complexity, a double-track model is implemented in~\cite{Double-friction}, where longitudinal and lateral load transfers are considered along with aerodynamic loads and Pacejka's Magic Formula~\cite{Pac}. 
In~\cite{Double-aero} and~\cite{Double-LSD} the double track model is further refined by considering four-wheel drive / active aerodinamic control and a limited-slip differential, respectively. Instead, Limebeer et al.~\cite{Double-F1} develop a double track vehicle model embedded in a 3D track. Hence, they take into account the effects of track geometry when computing load transfers and vehicle absolute velocity.

As the last stage of complexity, a multibody approach can be used to increase the level of details. In particular, in~\cite{Multibody-planar} a 2D multibody dynamic model is developed where the rear wheels are fixed to the chassis - making it a single rigid body - while the front wheels are independent bodies pinned to the main chassis via revolute joints.  Dal Bianco et al.~\cite{Multibody-3D} extended further and developed a 3D multibody car model with 14 degrees of freedom.

Even if successful, all the mentioned contributions do not provide a systematic framework for the assembly of the vehicle dynamic equations, especially when considering their motion on 3D tracks. Their approaches seem episodic lacking a systematic procedure. Moreover, they do not exploit the recent developments in recursive dynamics algorithms, quite popular, on the contrary, in the fields of robotics and general computational dynamics, see e.g.~\cite{Mueller2003} and~\cite{Featherstone}.

In this work, we try to fill this gap by presenting a unified framework to systematically build a vehicle model that balances model accuracy and efficiency. More specifically, looking at a vehicle as a serial robot, a Lie-group based race car model is developed. The effects of 3D track geometry are directly included with an original formulation since a generalized kinematic joint enables the natural embedding of the car model into the 3D track. The dynamics equations are obtained by merging and efficient recursive formulation based on the \emph{Articulated Body Algorithm} (ABA)~\cite{Featherstone} and a rigorous treatment of vehicle dynamics~\cite{Guiggiani}. 
Finally, proper algebraic equations allow to incorporate fundamental phenomena in vehicle dynamics such as lateral load transfers and nonlinear dependance of tire forces on vertical loads within the ABA formulation. A noteworthy result is that our framework opens up the possibility to directly employ efficient and open-source rigid body dynamics libraries (see, e.g.~\cite{RBDL} documented in~\cite{Felis2016}) also within the vehicle dynamics context.

The paper is organized as follows: Section~\ref{sec:kinematic} focuses on track and vehicle parametrization, highlighting the fundamental aspects such as the mathematical description of the track, reference frames, and kinematic chain that describes the vehicle structure.
In Sect.~\ref{sec:dynamic} the vehicle dynamic model is obtained through the ABA formulation. Here a reconciliation of tire forces and load transfers with the wrench introduced in the ABA setting is described. To this sake, the suspension constitutive algebraic equations in the framework proposed in~\cite{Guiggiani} are key.
Finally, Sect.~\ref{sec:application} shows some numerical results from the solution of a MLTP, that is a trajectory optimization problem on a sector of the N\"urburgring circuit. Here, the proposed framework has been compelling in building up efficient dynamic model equations.

\section{Kinematic Model}\label{sec:kinematic}

With reference to Fig.~\ref{fig:3dvehicle}, the kinematic model of a vehicle travelling on a 3D track is devised as a serial kinematic chain whose root node consists of a fixed Cartesian reference frame $\Bo$ and whose \emph{end-effector} represent the vehicle sprung mass, to which frame $\Bsix$ is attached. The serial chain starts with a \emph{complex joint} that accounts for advancing tangentially to the road along the track centerline and proceeds with virtual translational and revolute joints.
To efficiently parameterize the posture of the $i$-th body respect to the fixed reference frame $\Bo$, we employ the \emph{body-fixed reference frame (local)} version of the \emph{Product of Exponentials} (POE) formula~\cite{Muller}, i.e.
\begin{equation}
\label{eq:POE}
\gbobi(q) = \prod_{k = 1}^{i} \gbkmbk(0) e^{\hxk q_{k}}.
\end{equation}
Here $\gbobi \in \text{SE(3)}$ denotes the posture of  $\Bi$ with respect to  $\Bo$, $\gbkmbk(0)$ represents the initial configuration of $\Bk$ w.r.t. $\Bkmone$, $\hxk$ are the twists of the joints defining the kinematic chain, and $q = [q_1, \dots, q_i]^T$ are the exponential coordinates of the 2nd kind~\cite{Sastry} for a local representation of SE(3) for the $i$-th body.

Symbol $X_{k}$ is a shorthand for $\xbj{k}{k}$, i.e. $X_{k} = \xbj{k}{k}$ when expressed in the attached local frame $\Bk$, the right superscript denoting the reading frame $\Bk$. In the general case
\begin{equation}
	\xbj{k}{i} = \Ad_{\gbibj} \xbj{k}{j}
\end{equation}
where, the \emph{Adjoint} transformation $\Ad_{\gbibj}$ maps the same twist $\xbj{k}{\phantom{j}}$ from reading frame $\Bj$ to $\Bi$.

The rigid-body velocity $\hVbobi$ of $\Bi$ w.r.t. $\Bo$ in the moving frame $\Bi$ is given (as a 4x4 matrix) by the following formula

\begin{equation}
\label{eq:vel_twist}
	\hVbobi := \gbobi^{-1}\dotgbobi=
	\begin{bmatrix}
		\homegabobi & \vbobi  \\
		 0_{1\text{x}3} & 0 \\
	\end{bmatrix}
\end{equation}
where, given the 3x3 rotation matrix $\Rbobi$ from $\Bo$ to
$\Bi$, $\homegabobi := \Rbobi^{T}\dotRbobi$ is the skew-symmetric matrix
of the angular velocity components (in $\Bi$) of $\Bi$ w.r.t. $\Bo$, and $\vbobi =  \Rbobi^{T}\dotdbobi$ are the components (in $\Bi$) of the velocity of the origin $O_i$ with respect to $O_0$.
Equation~(\ref{eq:vel_twist}) can be rewritten (as a 6x1 vector) in a
convenient form by factoring out the joint velocities $\dot{q}$ as follows
\begin{equation}
\Vbobi = \Jbobi(q)\dot{q}
\end{equation}
where $q = [q_1 \cdots q_i]^T $ and the \emph{distal Jacobian} $\Jbobi$ can be computed as
\begin{equation}
	\Jbobi(q) = [\xd{1} \cdots \xd{i}], \hspace{5mm} \xd{k} = \Ad_{C_{k, i}^{-1}} \xbj{k}{},
\end{equation}
where we define $C_{k, i} = e^{\hxk \qk} g_{k,i}$ and $k = 1, \dots, i$.

Similarly to twist formulation, $\wbj{k}{k} \in \mathbb{R}^{6}$ denotes the components in $\Bk$ of the wrench exerted on the ${k}$-th body.
A generic wrench is transformed in a different frame, into an equivalent counterpart as follows
\begin{equation}
\label{eq:wrench_change}
\wbj{k}{i} =  \Ad_{\gbibj}^{*} \begin{bmatrix} f_{k}^{j}  \\ m_{k}^{j}\\\end{bmatrix} = \Ad_{\gbibj}^{*}  \wbj{k}{j}
\end{equation}
where, $f_k^j$ are the components of the force acting on body $k$, expressed in $\Bj$, and $m_{k}^{j}$ the components with respect to  $O_{j}$ and in $\Bj$ of the resulting moment applied to body $k$. The operator $\Ad_{g}^{*} = \Ad_{g}^{-T}$ maps the same wrench in different reading frames.

\subsection{Track Parametrization}

To build a 3D analytical model of the track centerline (spine), which is continuously differentiable and capable to efficiently represent complex shapes, while remaining numerically stable, 3D NURBS curves~\cite{nurbs} are employed. Analytically, the track spine curve $C(\al)$ is defined by the position vector $\bp(\al)$ such that
\begin{equation}
C(\al) = \{[\bp(\al)]^{G} = [p_x(\al)\, p_y(\al)\, p_z(\al)]^{T} \in \mathbb{R}^{3}:\al \in [0,1]\}.
\end{equation}
In our representation, $\al$ is not necessarily the curvilinear abscissa $s$ (arc length), but a generic curvilinear parameter. The relationship between $s$ and $\al$ is described by the following equation
\begin{equation}
\dfrac{\der s}{\der \al} = \dsda
\end{equation}
where $\bp_{,\al} = \der \bp/\der\al$.

\begin{figure}
	\centering
	\includegraphics[width=1\linewidth]{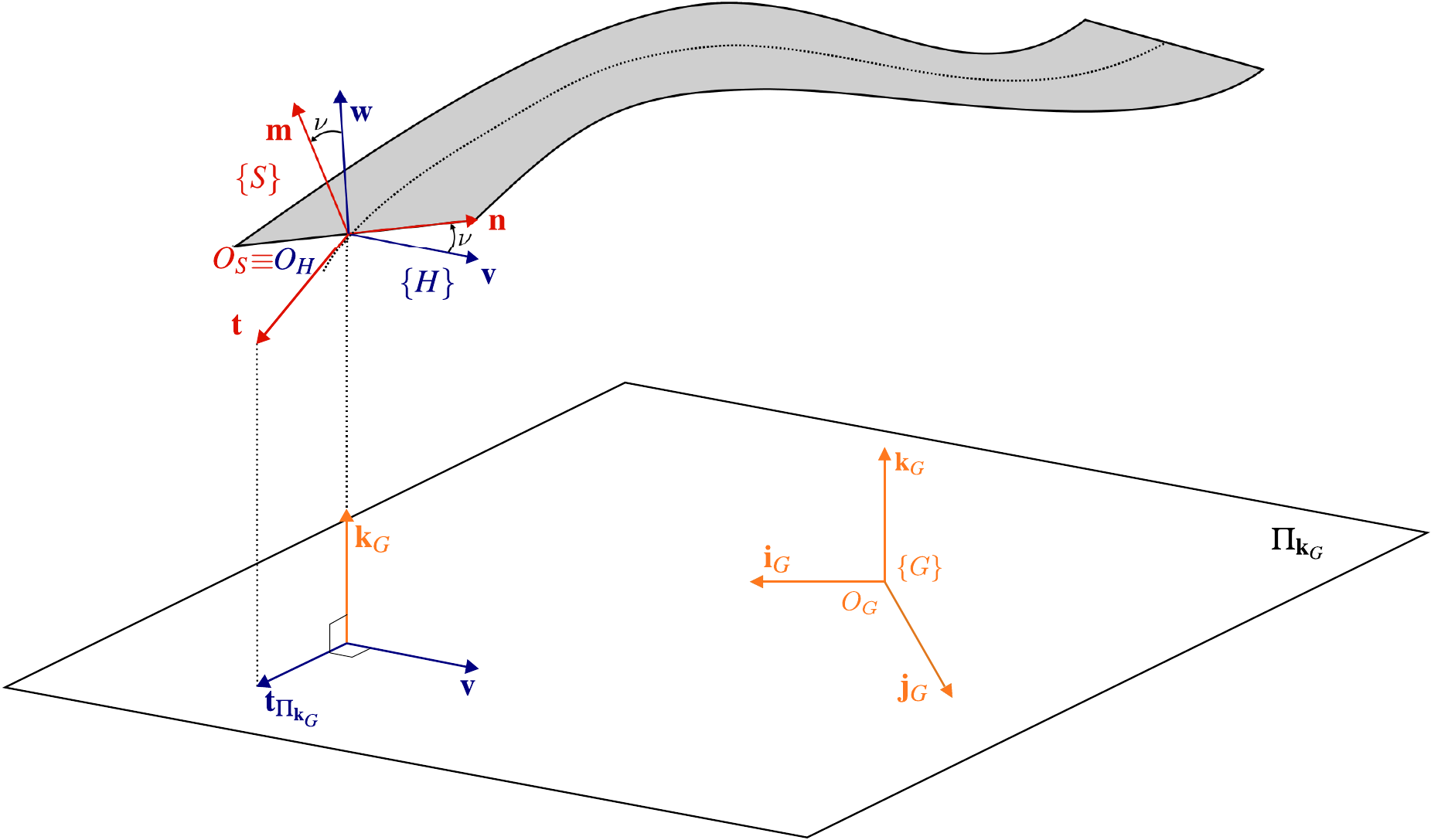}
	\caption{3D ribbon track with intermediate reference frame $\{H\}$ and track reference frame $\{S\}$.}
	\label{fig:track}
\end{figure}

In order to define precisely the 3D ribbon track frame $\Sf = (O_{S}; [\bt \,\, \bn \,\, \bm])$ that follows the track spine (see Fig.~\ref{fig:track}), an intermediate frame $\Hf = (O_{H}; [\bt \,\, \bv \,\, \bw])$ is introduced.
Here $\bt = \der\bp/\der s$ is the unit vector tangent to $C$, $\bv$ is the unit vector obtained normalizing $\mathbf{k}_{G}  \times \mathbf{t}_{\Pi_{\mathbf{k}_G}}$, where $\mathbf{k}_{G}$ is the unit vector representing the vertical direction of the ground-fixed reference frame and $\mathbf{t}_{\Pi_{\mathbf{k}_G}}$ is the projection of $\bt$ on the plane $\Pi_{\mathbf{k}_G}$, which is perpendicular to $\mathbf{k}_{G}$; finally $\bw$ is obtained as $\bt \times \bv$.
Then, $\Sf$ is obtained by rotating $\Hf$ about  $\bt$ through an angle $\nu$, which represents the track banking.

It is worth remarking that the complex track joint cannot be analyzed using the exponential approach (see~(\ref{eq:POE})).
Hence the transformation matrix $g_{G, S}$ and the rigid-body velocity $\Vgs$ of $\Sf$ w.r.t. $\Gf$ expressed in $\Sf$ are derived following the general definition~\cite{Sastry}. Once the track has been parametrized and the NURBS analytical model is available, the quantities $[\bt\, \bn\, \bm]$ can be computed and $g_{G, S}$ can be evaluated as
\begin{equation}
\label{eq:ggs}
g_{G, S}(\al) = \begin{bmatrix} R_{G, S}(\al) & C(\al)\\ 0_{1 \times 3} & 1 \end{bmatrix}; \hspace{5mm} R_{G, S} = [t^{G}\, n^{G}\, m^{G}]
\end{equation}

where $R_{G, S}$ is the rotation matrix from $\Gf$ to $\Sf$ and $t^{G}$ and $n^{G}$ and $m^{G}$ are the components of $\bt$, $\bn$, $\bm$ in the fixed-ground reference frame $\Gf$.

Instead the velocity $\Vgs$ can be computed as
\begin{equation}
\label{eq:track_twist}
\Vgs = \begin{bmatrix} v_{G, S}^{S} \\ \omega_{G, S}^{S} \end{bmatrix} = \begin{bmatrix} t^{S} \dot{s} \\ \Omega_{G, S}^{S} \dot{s} \end{bmatrix} = T_{G, S}^{S} \dot{s} = T_{G, S}^{S} \dot{\al} \dsda.
\end{equation}
Here, $T_{G, S}^{S}$ is the \emph{geometric twist} obtained by differentiation of $C$ and $R_{G, S}$ with respect to $s$, $t^{S}$ is the unit tangent vector to the centerline, and the geometric angular velocity $\Omega_{G, S}^{S}$ has its hat form defined by the following equation
\begin{equation}
\hat{\Omega}_{G, S}^{S} = R_{G, S}^{T} \frac{\der R_{G, S}}{\der s}.
\end{equation}

\subsection{Vehicle Parametrization}

\begin{figure}
	\centering
	\includegraphics[width=1\linewidth]{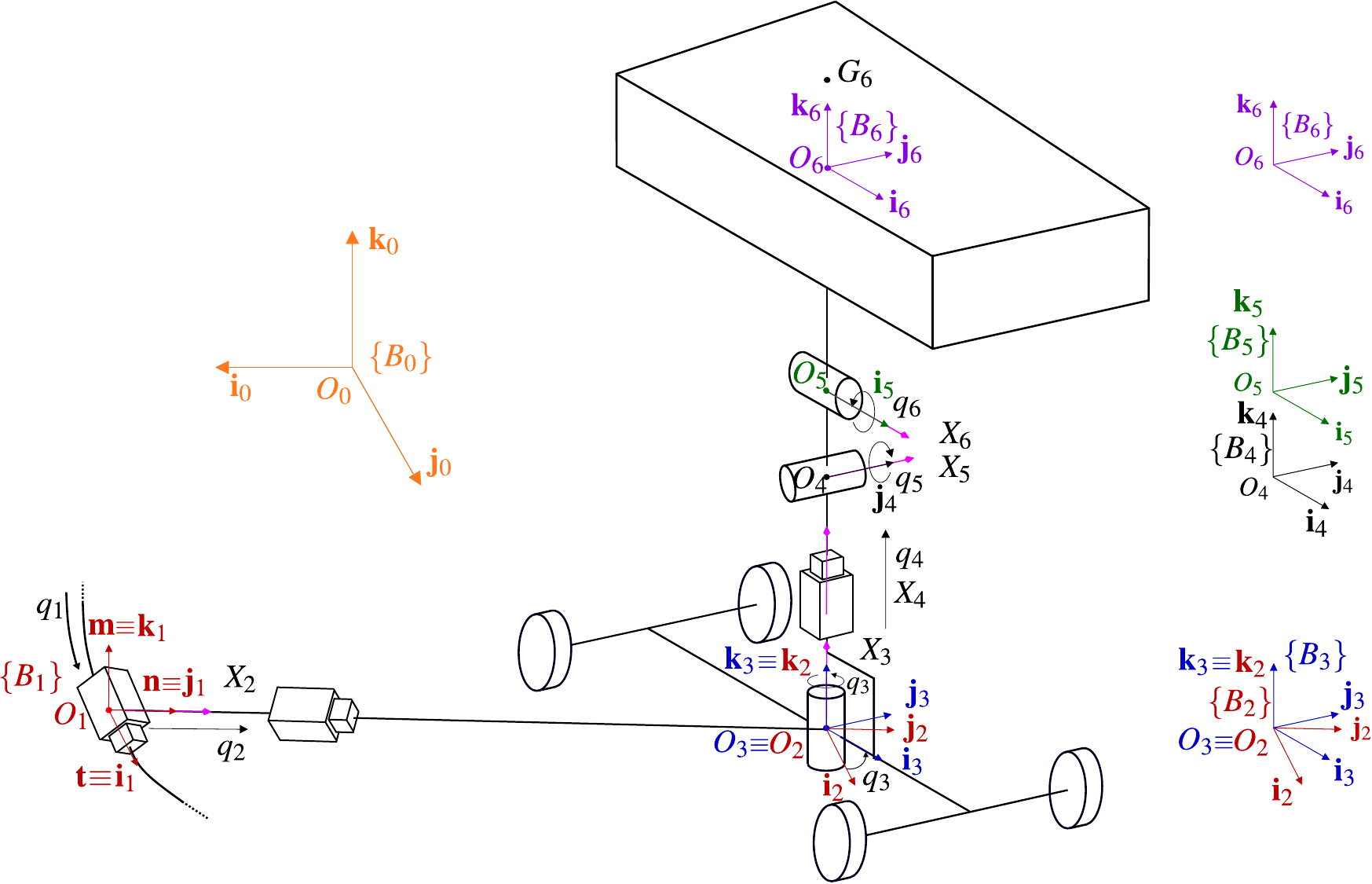}
	\caption{Kinematic chain of the 3D vehicle model with the reference frames and degrees of freedom described by coordinates $q$.}
	\label{fig:3dvehicle}
\end{figure}

The vehicle kinematic chain is shown in Fig.~\ref{fig:3dvehicle}. Here, the reference frames from $\Bo$ to $\Bsix$, kinematic joints along with their corresponding joint variables, and the associated twists are depicted.

Starting from the ground $\Bo$, the first joint is associated with the track and  transforms the ground frame $\Bo$ into the track frame $\Bone$. Its motion is parametrized by the $q_1$ coordinate.
Then, variables $q_2$ and $q_3$, associated with (virtual) prismatic and revolute joints respectively, encode the vehicle degrees of freedom w.r.t. to $\Bone$. Hence, the car can translate along the normal direction $\mathbf{j}_1$, defining the frame $\Btwo$, and rotate along the vertical direction $\mathbf{k}_2$, thus defining $\Bthree$.
In particular, this frame is located at the road level and it is fixed to the car axles plane where the interactions between road and vehicle are exchanged.

The remaining joints angles $q_4$, $q_5$ and $q_6$, parametrize the relative motions of the car body frame $\Bsix$ with respect to the car axles plane, due to the suspension system. In particular, $q_4$ is the vertical displacement, and $q_5$, $q_6$ are, according to common vehicle dynamics notation~\cite{Guiggiani}, the pitch and roll angles, respectively.

It is worth observing that the last two revolute joints have intersecting axes. Furthermore, our reference frames definition implies that $O_4 \equiv O_5 \equiv O_6$. In particular, point $O_6$ does not coincide with car body center of mass $G_6$ (which is located above along $\mathbf{k}_6$ direction), but coincides with the \emph{vehicle invariant point} (VIP)~\cite{Guiggiani}. This point, regardless of the roll angle, remains centered with respect to the four contact patches, hence in the middle of the vehicle, even when it rolls. This property makes such point the best option to monitor the vehicle position.

While the joints from $\Bone$ to $\Bsix$ are well characterized by their twists, and can be parameterized conveniently by the exponential approach, the track joint (the first one), requires a specific formulation, as detailed in the previous subsection.
Considering that $\Sf$ and $\Gf$ frames, introduced in Fig.~\ref{fig:track}, become $\{B_{0}\}$ and $\{B_{1}\}$ according to the notation of Fig.~\ref{fig:3dvehicle}, and that the joint variable $\al$ becomes $q_1$, we can rewrite~\eqref{eq:ggs} and~\eqref{eq:track_twist} as follows
\begin{subequations}
	\begin{align}
	\label{eq:g01}
	& \gbobone = \begin{bmatrix} R_{0, 1}(q_1) & C(q_1)\\ 0_{1 \times 3} & 1 \end{bmatrix}  \\
	\label{eq:V01}
	& \Voone = T_{0, 1}^{1} \dsdqone  \dot{q_{1}} = \Joone \dot{q}_{1}.
	\end{align}
\end{subequations}

\section{Dynamic Model}\label{sec:dynamic}

Once the vehicle has been parametrized by means of the Lie Group machinery, the equations of motion can be derived systematically.
To this end the \emph{Articulated Body Algorithm} (ABA)~\cite{Featherstone} is employed for reasons discussed later.

The dynamics of a generic body $k$ connected to a parent joint can be written through the Newton-Euler equations
\begin{equation}
\label{eq:motioni}
\WS = \Mk\dVk + \bk,
\end{equation}
where $\WS$ is the  wrench exerted on body $k$ \emph{through} the previous connection joint, $\Mk$ is the inertia matrix $k$, $\dVk$\footnote{The subscript $0$ is omitted when referring to the motion w.r.t. the ground.} is the rigid-body acceleration, and $\bk$ is the bias force computed as

\begin{equation}
\label{eq:bias}
\bk = \sadvi \Mk\Vk - \WE.
\end{equation}
In~\eqref{eq:bias}, the first term represents the generalized gyroscopic forces/torques, which are bilinear in $\Vk$, while $\WE$ is the wrench exerted by the forces \emph{directly} applied to body $k$. The mathematical operator $\ad_{V}$ in~\eqref{eq:bias} transforms the input vector $V$ in a $6 \times 6$ matrix as follows
\begin{equation}
\ad_{V} = \begin{bmatrix}
\hat{\omega} & \hat{v} \\ 0_{3 \times 3} & \hat{\omega},
\end{bmatrix}
\end{equation}
and serves to compute the Lie derivative between two vector fields.
Referring to~\eqref{eq:bias}, it is worth recalling that $\ad_{V}^* = -\ad_{V}^{T}$.

Key of the ABA algorithm is the concept of \emph{articulated body}, defined as a collection of $N_{B}$ rigid bodies connected by movable joints (active and/or passive). Remarkably, if $k$ is the \emph{handle} (first body) of an articulated body, its dynamics can still be written using~\eqref{eq:motioni} in the following form
\begin{equation}
\label{eq:dyn_hat}
\WS = \hMk\dVk + \hbk.
\end{equation}
Here $\hMk$ and $\hbk$ are the generalized inertial and bias terms, which account for inertia and bias forces of the children bodies in the kinematic chain which are structurally transmitted backwards to the handle $k$.

The explicit expressions of the articulated body inertia and bias terms $\hMk$ and $\hbk$, respectively, along with other fundamental aspects of the ABA algorithm are given in the next subsection.

\subsection{Forward Dynamics via a Tailored ABA Formulation}
The \emph{Articulated Body Algorithm} consists of three different steps.

\subsubsection{Forward Propagation of Posture and Velocity}
In this step, starting from the handle body, rigid-body postures and velocities are being propagated from the ground to the car body.


\begin{algorithm}
		\floatname{algorithm}{Step}
 	\caption{Forward propagation of postures and velocities }\label{alg:step1}
 	\begin{algorithmic}[1]
 		\For {$k = 1$ to $N_B$}
 		\If{$k = 1$}						\Comment{Track Joint}
		\State $\gbobone = \gbobone(q_1)$ \Comment{\eqref{eq:g01}}
 		\State $V_1^1 = J_1^1 \dot{q}_{1}$   \Comment{\eqref{eq:V01}}
 		\Else
 		\State $\glambda(q_{k}) = \glambda(0)e^{\hxk q_{k}}$ \Comment{Posture}
 		\vspace{1mm}
 		\State $\Vk = \mAdlambda\Vlambda + X_k\dot{q}_k$     \Comment{Velocity}
 		\EndIf
 		\EndFor
 	\end{algorithmic}
\end{algorithm}

The number of rigid bodies of our articulated body is $N_B = 6$. These are identified by frames $\Bone,\ldots, \Bsix$ and their inertial properties are introduced in the next ABA step. As detailed in Section~\ref{sec:kinematic}, it is worth noting that the first joint (track transformation) is treated separately, via the homogeneous matrix $\gbobone$  and the Jacobian $J_1^1$.

\subsubsection{Evaluation of the Generalized Bias Force and Articulated Body Inertia}
 In this step, starting from the distal body (the last body of the kinematic chain) we evaluate $\hMk$ and $\hbk$ introduced in~\eqref{eq:dyn_hat} for a generic body $k$.

\begin{algorithm}
	\floatname{algorithm}{Step}
	\caption{Evaluation of generalized bias force and articulated body inertia}\label{alg:step2}
	\begin{algorithmic}[1]
		\For {$k = N_B$ to $1$}
		\If {$k = N_B$}
		\vspace{1mm}
		\State $\hMk = \Mk$ \Comment{Articulated body inertia}
		\State $\hbk = \bk$ \Comment{Generalized bias force}
		\Else

		\State $l = k+1$
				\vspace{1mm}
		\State $\hMk = \Mk +  A_{k l}^{*} \bMl A_{l k} $
		\vspace{1mm}
		\State $\hbk = \bk + \bbl$
		\EndIf
		\EndFor
	\end{algorithmic}
\end{algorithm}

The quantities $\bMl$ and $\bbl$ are calculated as
\begin{subequations}
	\begin{align}
	&   \bMl = \hMl - \dfrac{\hMl X_l X_l^{T} \hMl}{X_l^{T} \hMl X_l} \\
	& \bbl = \Bigg[A_{k l}^{*} \hbl - A_{k l}^{*} \bMl \adl \Vkl + \dfrac{A_{k l}^{*} \hMl X_l (\tau_l - X_l^{T}\hbl)}{X_l^{T}\hMl X_l} \Bigg],
	\end{align}
\end{subequations}
where the shorthand notation $A_{l k} = \Alambda$ is used and $\tau_l$ is the active joint force or torque, depending on the joint nature. In the vehicle model here proposed joints are not actuated and non-zero $\tau_l$'s are only generated by springs and dampers.

{\bf Step 2} can be easily implemented once the terms $\Mk$, $\tau_l$ and $\WEk$ have been defined for each body.

In our serial kinematic chain, only the inertias of sprung $M_{6}^{6}$ and unsprung masses $M_{3}^{3}$, as usual in vehicle dynamics~\cite{Guiggiani}, are different from zero.

Regarding the active joint force/torque $\tau_l$, we clearly distinguish the first three joints, obviously fictitious (thus passive), from the last ones. Therefore, while $\tau_1 = \tau_2 = \tau_3 = 0$, the last ones $\tau_4, \tau_5$ and $\tau_6$, even if not actuated, can develop a force/torque due to the presence of springs and dampers. Their constitutive equations are described by
\begin{subequations}
	\begin{align}
	&  \tau_6 = -k_{\phi}q_6 -c_{\phi}\dot{q}_6  \\
	&  \tau_5 = -k_{\theta}q_5 -c_{\theta}\dot{q}_5  \\
	&  \tau_4 = -k\,q_4 -c\,\dot{q}_4,
	\end{align}
\end{subequations}
where $k_{\phi}$, $k_{\theta}$ and $k$ are first-order approximations of the equivalent roll, pitch and vertical stiffness, respectively; instead $c_{\phi}$, $c_{\theta}$ and $c$ are the equivalent roll, pitch, and vertical damping, respectively. Employing symbol $p$ to represent either $k$ or $c$, their explicit expressions can be computed as follows
\begin{subequations}
	\begin{align}
	&  p_{\phi}  = \dfrac{p_{11} + p_{12} }{4}t_1^2 + \dfrac{p_{21} + p_{22} }{4}t_2^2   \\
	&  p_{\theta}  = (p_{11} + p_{12})a_1^2 +(p_{21} + p_{22})a_2^2  \\
	&  p = p_{11} + p_{12} + p_{21} + p_{22}
	\end{align}
\end{subequations}
where, according to the notation in~\cite{Guiggiani}, the first subscript of $p_{ij}$ refers to the axle ($i = 1;2$ front, rear), whereas $j$ refers to the vehicle side ($j = 1;2$ left, right). As usual, $t_1$ and $t_2$ are the front and rear tracks of the car.

Finally, to evaluate $\bk$ as in~\eqref{eq:bias}, the external wrenches have to be defined.
The only contributions come from the aerodynamic forces, applied to the car body (fixed to $\Bsix$), and the interaction between the axle body (fixed to $\Bthree$) and the road, whereas gravity contribution is treated separately, as explained in {\bf Step 3}. As for the aerodynamic wrench $W_{6E}^{6}$, it is convenient to evaluate it in $\Bthree$ and then express it back in $\Bsix$ through~\eqref{eq:wrench_change} to model also its effects on roll, pitch and bounce motions. Therefore, its expression is computed as $\WEaero = \sAdts \WEaerot$, where
\begin{equation}
	\begin{aligned}
	\WEaerot &= [f_{x_a}, 0, f_{z_a}, 0, m_{y_a}, 0]^{T}\\
	&=-\dfrac{1}{2} \rho S (v_{3_{x}}^3)^{2} [C_{x}, 0, C_{z}, 0, C_{z2}a_2 - 	C_{z1}a_1, 0]^{T}.
	\end{aligned}
\end{equation}

Here, $\rho$ is the air density, $S$ is the vehicle frontal area, $v_{3_{x}}^3$ is the $v_{3}^3$ component along $\mathbf{i}_3$, and $a_1$, $a_2$ are the longitudinal distances of $G_6$ from the front and rear axles, respectively. The drag coefficient is $C_x > 0$, the downforce coefficient is $C_z > 0$ and the front and rear downforce coefficients $C_{z2}$, $C_{z1}$ are such that $C_z = C_{z1} + C_{z2}$.

\begin{figure}
	\centering
	\includegraphics[width=1\linewidth]{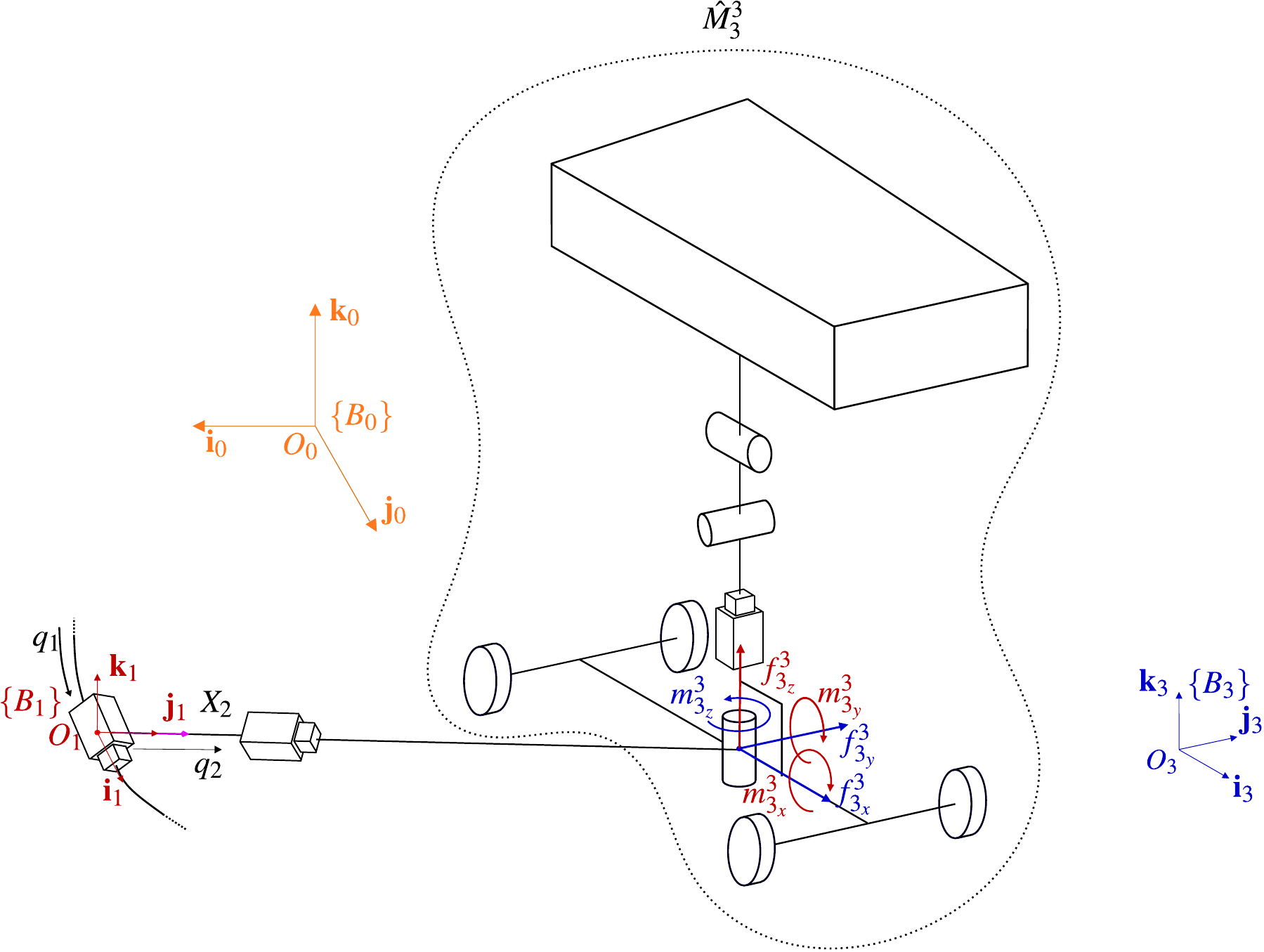}
	\caption{Step 2 of the \emph{Articulated Body Algorithm}: Evaluation of \emph{Articulated body inertia} $\hat{M}_3^{3}$ and representation of \emph{in-plane} (blue) and \emph{out-of-plane} (red) wrenches.}
	\label{fig:3dvehicleaba}
\end{figure}

The other non-zero external wrench $W_{3_E}^{3}$ is applied directly on the axle body (fixed to $\Bthree$) and accounts for \emph{a portion} of the interactions between road and vehicle.
In a real vehicle, and also in our model, the totality of the external forces that act on the car, except for the aerodynamic ones, are developed through the contact between tires and road.
However, focusing on body $\Bthree$ and considering it as the handle of an articulated body (from $\Bthree$ to $\Bsix$) to which the ABA machinery is applied, it is more convenient to encode \emph{in-plane} components in the external wrench $W_{3_{E}}^3$ and \emph{out-of-plane} components in the structural wrench $W_{3_{J}}^3$, as shown in Fig.~\ref{fig:3dvehicleaba}.
Therefore, the above perspective suggests the following partition
\begin{equation}
W_{3}^{3} = W_{3_J}^{3} + W_{3_E}^{3} ,
\end{equation}
where $W^3_3$ represents the global system of forces and torques generated at the four contact patches between road and tires.
More in details we define
\begin{subequations}
	\begin{align}
	&W_{3_{J}}^3 = [0, 0, f_{3_z}^{3}, m_{3_x}^{3}, m_{3_y}^{3}, 0]^{T}\\
	&W_{3_{E}}^3 = [f_{3_x}^{3}, f_{3_y}^{3}, 0, 0, 0, m_{3_z}^{3}]^{T}\label{eq:W33E}.
	\end{align}
\end{subequations}

Considering that the first three joints are passive, $W_{3_{J}}^3$ represents the structural wrench. It is worth remarking that its non-zero components $f_{3_z}^{3}$, $m_{3_x}^{3}$ and $m_{3_y}^{3}$ are the \emph{out-of-plane} force and torques that can be thought, in the ABA perspective, as those structurally absorbed by the first three joints of the virtual kinematic chain. These components restrain $\Bthree$ to stay on the track. Instead, \emph{in-plane} (plane locally tangent to the road) components $f_{3_x}^{3}$, $f_{3_y}^{3}$ and $m_{3_z}^{3}$ are treated as external forces which account for the vehicle traction and are embedded in $W^3_{3_E}$. These will be linked, in the next subsection, to the control inputs of our model.

\subsubsection{Forward Propagation of Acceleration}
In this step, starting from the first body, we compute and propagate joint accelerations $\ddot{q}_k$ to obtain the rigid-body accelerations $\dVk$.
\begin{algorithm}
	\floatname{algorithm}{Step}
	\caption{Forward Propagation of Joint Acceleration}\label{alg:step3}
	\begin{algorithmic}[1]
		\For {$k = 1$ to $N_B$}
		\If {$k = 1$}
		\vspace{2mm}
		\State $\ddot{q}_{1} = -\dfrac{(J_1^1)^{T}[\hat{M}_1^1(\dot{V}_{0}^1 + J_{1,q_1}^1\dot{q_1}^2)+\hat{b}_1^1]}{(J_1^1)^{T}\hat{M}_1^1 (J_1^1)}$
		\vspace{2mm}
		\State $\dot{V}_1^1 = \Ad_{g_{1,0}}\dot{V}_{0}^0 + J_{1,q_1}^1\dot{q_1}^2 + J_1^1\ddot{q}_1$
		\Else
		\State	$\lambda = k-1$
		\State $\ddot{q}_k = \dfrac{\tau_k - X_k^{T}[\hMk(\dot{V}_{\lambda}^{k} - \adk V_{\lambda}^{k}) + \hbk]}{X_k^{T}\hMk X_k}$
		\vspace{1mm}
		\State $\dVk = A_{k\lambda} \dVlk + X_k \ddot{q}_k - \adk  A_{k\lambda} \Vlambda $
		\EndIf
		\EndFor
	\end{algorithmic}
\end{algorithm}

This procedure is presented  in the pseudo-code \textbf{Step~\ref{alg:step3}}, where $J_{1,q_1}^1 = \der J_1^1/\der q_1$.

As in \textbf{Step 1}, the first joint is treated separately, due to its non-standard nature. Furthermore, in order to model the presence of gravity, we introduce a fictitious acceleration on $\Bo$ (which is automatically propagated through the kinematic chain) by posing
\begin{subequations}
	\begin{align}
	&\dot{V}_0^0 = [0, 0, a_g, 0, 0, 0]^{T}\\
	&V_0^0 = [0, 0, 0, 0, 0, 0]^{T},
	\end{align}
\end{subequations}
where $a_g = 9.81 $ m/s$^2$ is the gravity acceleration.

After \textbf{Step~\ref{alg:step3}}, having computed $\dot{V}_3^3$,  we can calculate the structural wrench $W_{3_J}^{3}$ through~\eqref{eq:dyn_hat} as follows
\begin{equation}
\label{eq:dyn_3}
W_{3_J}^{3} = \hat{M}_{3}^{3}\dot{V}_{3}^{3} + \hat{b}_{3}^{3}.
\end{equation}
Considering that $W_{3_J}^{3}$ has \emph{only three non-zero components},~\eqref{eq:dyn_3} represent three equations linking $f_{3_z}^{3}$, $m_{3_x}^{3}$ and $m_{3_y}^{3}$ to the inertial, bias and acceleration terms obtained through the ABA algorithm. More in detail, taking into account that the $\tau_k$'s depend only on $q$ and $\dot{q}$, and that bias $\hat{b}_3^3$ depends on $W_{3_E}^3$, the following dependencies hold
\begin{equation}
\label{eq:W3jf}
W_{3_J}^{3} = W_{3_J}^{3}(q, \dot{q}, \ddot{q}, W_{3_E}^3)
\end{equation}

\subsection{Reconciliation of ABA Wrenches with Tire Forces and Load Transfers}
The paramount aspect that characterizes vehicle dynamics is the interaction between road and tires. As explained in the previous subsection, in our model this interaction is encoded in wrenches $W_{3_J}^{3}$ and $W_{3_E}^{3}$. Therefore, in order to model the dynamics of an actual vehicle with tires, it is necessary to link them to the actual forces exchanged within the four contact patches.

The generic wrench of the $ij$-th ($i$, again, refers to the axle and $j$ refers to left/right sides) wheel is composed only by the three components $f_{ij_x}$, $f_{ij_y}$, $f_{ij_z}$, which are expressed in the corresponding $\{B_{ij}\}$ frame\footnote{Each $\{B_{ij}\}$ has its origin in the center of the contact patch of the $ij$-th wheel and it is rotated w.r.t. $\{B_3\}$ of the wheel steering angle $\delta_{ij}$ about $\mathbf{k}_{ij}$.}. Since we consider a vehicle with only front wheel steering and with a parallel steering law, we pose $\delta_{21} = \delta_{22} = 0$ and $\delta_{11} = \delta_{12} = \delta$.

The first force we analyze is the vertical one. Inspired by~\cite{Guiggiani} we can write
\begin{equation}
\label{eq:fz}
f_{ij_z} = f_{z_{i0}} + f_{z_{ia}} + \Delta f_z + (-1)^j \Delta f_{z_i},
\end{equation}
where $f_{z_{i0}}$ is the static load, $f_{z_{ia}}$ is the aerodynamic force, and $\Delta f_z$, $\Delta f_{z_i}$ are the longitudinal and lateral load transfers, respectively. Equations~\eqref{eq:fz} (one for each wheel) represent implicit equations in the $f_{ij_z}$ terms. To clarify this aspect, the four components of~\eqref{eq:fz} are analyzed, highlighting  their dependencies on $f_{ij_z}$ and on the non-zero components of $W_{3_J}^3$.

 By definition, the first component $f_{z_{i0}}$ is evaluated from its dynamic counterpart $f_{3_z}^3$ filtering out the aerodynamic force as follows
\begin{equation}
\label{eq:fzi0}
f_{z_{i0}}=f_{z_{i0}}(f_{3_z}^3) = (f_{3_z}^3 - f_{z_a})\dfrac{(l-a_i)}{2l},
\end{equation}
where $l = a_1 + a_2$ is the wheelbase. Then, the downforce is reintroduced via
\begin{equation}
\label{eq:fzia}
f_{z_{ia}} = \dfrac{1}{4}\rho C_{zi} S (v_{3_x}^3)^2.
\end{equation}
The longitudinal load transfer is obtained as
\begin{equation}
\label{eq:dfz}
\Delta f_z = \Delta f_z(m_{3_y}^3) = -(m_{3_y}^3 - m_{y_a})/(2l),
\end{equation}
deducting the aerodynamic moment since it has already been considered through $f_{z_{ia}}$ at each wheel.
Finally, according to~\cite[p.~152]{Guiggiani}, and assuming the tires to be perfectly rigid in the vertical direction, we can compute the lateral load transfers as follows
\begin{equation}
		\begin{aligned}
		\label{eq:dzi}
		\Delta f_{z_i} &= \Delta f_{z_i}(f_{11_z}, f_{12_z}, f_{21_z}, f_{22_z}) \\
		 &=  {\dfrac{k_{\phi_i}}{k_{\phi}t_i}[- m_{3_x}^{3} - (Y_1 h_{q_1} + Y_2 h_{q_2})]} + {\dfrac{Y_i h_{q_i}}{t_i}}.
		\end{aligned}
\end{equation}
Here, $Y_i = Y_i(f_{i1_z}, f_{i2_z})$ is the lateral force acting on $i$-th axle, expressed in $\Bthree$, $k_{\phi_i}$\footnote{$k_\phi=k_{\phi_1}+k_{\phi_2}$.} is the roll stiffness of the $i$-th axle and $h_{q_i}$ is the distance of the \emph{no-roll center} of the $i$-th suspension from the road~\cite[p.119]{Guiggiani}.

The explicit expressions of $Y_1$ and $Y_2$ are given by
\begin{subequations}
	\label{eq:Yi}
	\begin{align}
	    &\begin{aligned}
		Y_{1} &= Y_{1}(f_{11_z}, f_{12_z}) \\
		&= (f_{11_y} + f_{12_y})\cos(\delta) + (f_{11_x} + f_{12_x})\sin(\delta)
	    \end{aligned}\\
		&Y_{2} = Y_{2}(f_{21_z}, f_{22_z}) = f_{21_y} + f_{22_y}.
	\end{align}
\end{subequations}
These highlight the dependencies on the vertical forces of the lateral force $f_{ij_y}$ for each wheel which come from the tire model we adopt.

To describe the tire behaviour in the lateral direction we employ Pacejka's Magic Formula~\cite{Pac}, which reads
\begin{equation}
\label{eq:fy}
\begin{aligned}
f_{ij_{y}} &= f_{ij_y}(f_{ij_z}) \\
&= D_y \sin(C_y \arctan(B_y \al_{ij} - E_y(B_y \al_{ij} - \arctan(B_y \al_{ij})))).
\end{aligned}
\end{equation}
It is worth stressing that tire factors $D_y(f_{ij_z})$, $C_y(f_{ij_z})$, $B_y(f_{ij_z})$ and $E_y(f_{ij_z})$ depend on the characteristics of the tire and also explicitly on the vertical load $f_{ij_z}$.
The $\al_{ij}$'s are the tire slip angles which are assumed here almost equal for wheels of the same axle~\cite{Guiggiani}, although not strictly necessary. Their expressions are given by
\begin{subequations}
	\label{eq:slip}
	\begin{align}
	&\al_{11} = \al_{12} = \delta - \dfrac{v_{3_y}^{3} + \omega_{3_z}^{3}a_1}{v_{3_x}^{3}} \\
	&\al_{21} = \al_{22} = - \dfrac{v_{3_y}^{3} - \omega_{3_z}^{3}a_2}{v_{3_x}^{3}}
	\end{align}
\end{subequations}

Finally, considering that the vehicle here studied is rear wheel drive, equipped with an open differential (i.e. $f_{21_x} = f_{22_x}$), the longitudinal forces are given by the following equations
\begin{subequations}
	\label{eq:fx}
	\begin{align}
	&f_{11_x} = f_{12_x} = \dfrac{1}{2}f_{xb}k_b \\
	&f_{21_x} = f_{22_x} = \dfrac{1}{2}f_{xb}(1-k_b) + \dfrac{1}{2}f_{xa}
	\end{align}
\end{subequations}
where $k_b$ is the braking ratio, and $f_{xb}$, $f_{xa}$ have been introduced as the total \emph{braking} and \emph{traction} forces, respectively.

At this point it is important to underline how to combine the above equations in order to characterize the implicit equations given by~\eqref{eq:fz}.

Substituting~\eqref{eq:fy} in~\eqref{eq:Yi} and inserting the result in~\eqref{eq:dzi}, we obtain the explicit expression that links each $\Delta f_{z_i}$ to all four vertical loads $f_{{ij}_z}$.

At this point, equations \eqref{eq:dzi}, \eqref{eq:dfz}, \eqref{eq:fzia} and~\eqref{eq:fzi0} can be substituted in~\eqref{eq:fz}. It is worth noting that \eqref{eq:dzi}, \eqref{eq:dfz}, \eqref{eq:fzia} and~\eqref{eq:fzi0}, beside $v^3_{3x}$, contain $W_{3_J}^{3}$ components $m^3_{3_x}$, $m^3_{3_y}$ and $f^3_{3_z}$. However, according to~\eqref{eq:W3jf} and the results from \textbf{Step 1}, these can be eliminated in favor of $q$, $\dot{q}$ and $W_{3_E}^{3}$ components.

On the other side, since~\eqref{eq:dzi} contains $Y_i$ which depends through~\eqref{eq:Yi}, \eqref{eq:slip} and~\eqref{eq:fx} on $q$, $\dot{q}$, $f_{xa}$, $f_{xb}$ and $\delta$, equation~\eqref{eq:fz} becomes the following four implicit equations
\begin{equation}
\label{eq:fijzf}
f_{ij_z} = \widetilde{f}_{ij_z}(q, \dot{q}, \ddot{q}, W_{3_E}^{3}, f_{11_z}, f_{12_z}, f_{21_z}, f_{22_z}, f_{xa}, f_{xb}, \delta).
\end{equation}


It is worth noting that the system of vertical forces thus obtained is equivalent to $W_{3_{J}}^{3}$.

The final consistency condition requires the external wrench $W_{3_{E}}^{3}$, defined in~\eqref{eq:W33E} and appearing in~\eqref{eq:fijzf}, to be the resultant of \emph{in-plane} force components from tires as follows
\begin{subequations}
	\label{eq:congr}
	\begin{align}
	&\begin{aligned}
	f_{3_x}^{3} &= f_{3_x}^{3}(f_{11_z}, f_{12_z})\\
	&= X_1(f_{11_z}, f_{12_z}) + X_2
	\end{aligned}\\
	&\begin{aligned}
	f_{3_y}^{3} &= f_{3_y}^{3}(f_{11_z}, f_{12_z}, f_{21_z}, f_{22_z})\\
	&= Y_1(f_{11_z}, f_{12_z}) + Y_2(f_{21_z}, f_{22_z})
	\end{aligned}\\
	&\begin{aligned}
    m_{3_z}^{3} &= m_{3_z}^{3}(f_{11_z}, f_{12_z}, f_{21_z}, f_{22_z})\\
    &= Y_1(f_{11_z}, f_{12_z}) a_1 - Y_2(f_{21_z}, f_{22_z}) a_2
	\end{aligned}
	\end{align}
\end{subequations}
where, to compact the expression, $X_1(f_{11_z}, f_{12_z}) = (f_{11_x} + f_{12_x})\cos(\delta) - (f_{11_y} + f_{12_y})\sin(\delta)$ and $X_2 = f_{21_x} + f_{22_x}$.
Highlighting the whole dependencies contained in \eqref{eq:congr} we can write
\begin{equation}
\label{eq:W3ef}
W_{3_E}^3 = W_{3_E}^3(q, \dot{q}, f_{11_z}, f_{12_z}, f_{21_z}, f_{22_z}, f_{xa}, f_{xb}, \delta).
\end{equation}

Inserting~\eqref{eq:W3ef} in~\eqref{eq:fijzf} we can arrive at the following four expressions ($i,j=1,2$)
\begin{equation}
\label{eq:fijzf2}
f_{ij_z} = \widehat{f}_{ij_z}(q, \dot{q}, \ddot{q}, f_{11_z}, f_{12_z}, f_{21_z}, f_{22_z}, f_{xa}, f_{xb}, \delta).
\end{equation}
The dependency of $f_{ij_z}$ on $\ddot{q}$, leads to an implicit dynamic equation (see line 7 of \textbf{Step~\ref{alg:step3}}), due to the dependency of $W_{3_E}^3$ on $f_{ij_z}$. To cut open the resulting algebraic loop and restore the explicit form for the dynamic equations, in our implementation we introduce 7 algebraic variables as placeholders: the three non-zero component of $W_{3_E}^3$ and the four components $f_{ij_z}$ ($i,j=1,2$). Accordingly, we implement~\eqref{eq:congr} as three and~\eqref{eq:fijzf} as four algebraic equations. The resulting system, comprising the six ODEs coming from line 7 of \textbf{Step~\ref{alg:step3}}, becomes a DAE (Differential Algebraic Equations) system and can be approached, in the MLTP formulation, by introducing these equations as path equality constraints (see Section~\ref{sec:application}).

\section{Application To Trajectory Optimization}\label{sec:application}

The approach proposed is showcased in setting up the model equations of a Minimum Lap-Time Problem (MLTP). Here, the final goal is to find the optimal control inputs and the optimal trajectory that minimize the lap time of the vehicle on a given track.

In general MLTP can be formulated in time or spatial domains~(\cite{Lot-time} and \cite{Lot-spatial}, respectively). The first approach parameterizes the vehicle with its center of mass position with respect to the ground-fixed reference frame and time is the independent variable of the equations of motion. Instead, in the second approach, the vehicle position is described in terms of road coordinates and the curvilinear parameter (here $q_1$) of the track centerline is employed as the independent variable.

Since in our model the vehicle position and orientation are parameterized through the track coordinates ($q_1$, $q_2$ and $q_3$), the natural choice is to use the second approach. To this sake, the model equations obtained through the ABA algorithm have to be translated into the spatial domain.

Our model is characterized by the state vector $x = [q_1, q_2, q_3, q_4, q_5, q_6, \dot{q}_1, \dot{q}_2, \dot{q}_3, \dot{q}_4, \dot{q}_5, \dot{q}_6]$, by the control inputs $u = [f_{xa}, f_{xb}, \delta]$ and by the algebraic variables $z = [f_{11_z}, f_{12_z}, f_{21_z}, f_{22_z}, f_{3_x}^3, f_{3_y}^3, m_{3_z}^3]$.

The spatial formulation of the vehicle model is obtained computing $x_{,q_1} = \der x/\der q_{1}$, where $q_1$ is the track curvilinear parameter defined in Section~\ref{sec:kinematic}.
Hence, we can evaluate $x_{,q_1}$ as follows
\begin{equation}
\label{eq:ode}
x_{,q_1}(q_1) = \dot{x}/\dot{q}_1 = F(x(q_1), u(q_1), z(q_1))/\dot{q}_1
\end{equation}
where $F(\cdot)$ is the dynamic vector field, in which the $\ddot{q}_{i}$ components are obtained through the \emph{Articulated Body Algorithm}, and the system evolution is expressed as a function of $q_1$ instead of $t$.

\subsection{Formulation via Direct Collocation}

Among the many techniques that can be employed to solve OCPs~\cite{Betts}, the \emph{direct collocation method} is used in this study.
Its peculiarity is that the original OCP is discretized leading to a large (but sparse) \emph{Nonlinear Program} (NLP). The generic form of the resulting NLP is
\begin{equation}\label{eq:nlp}
\begin{split}
&\underset{x, v, u, z}{\text{minimize}}\hspace{10mm} \sum_{i = 0}^{N-1} l_{i}(x_i, v_i, u_i, z_i) + E(x_{N})\\
&\begin{split}
\text{subject to}\hspace{10mm} &g(x_{i},x_{i+1}, v_i, u_{i}, z_{i}) = 0,\\
&h(x_{i},u_{i},z_{i}) \leq 0,\\
& (i = 0, 1, \dots, N-1)\\
\end{split}\\
&\hspace{24mm} r(x_{N},u_{N-1},z_{N-1}) \leq 0.
\end{split}
\end{equation}

\begin{figure*}
	\centering
	\includegraphics[trim={0 0cm 0 0},clip,width=1\linewidth]{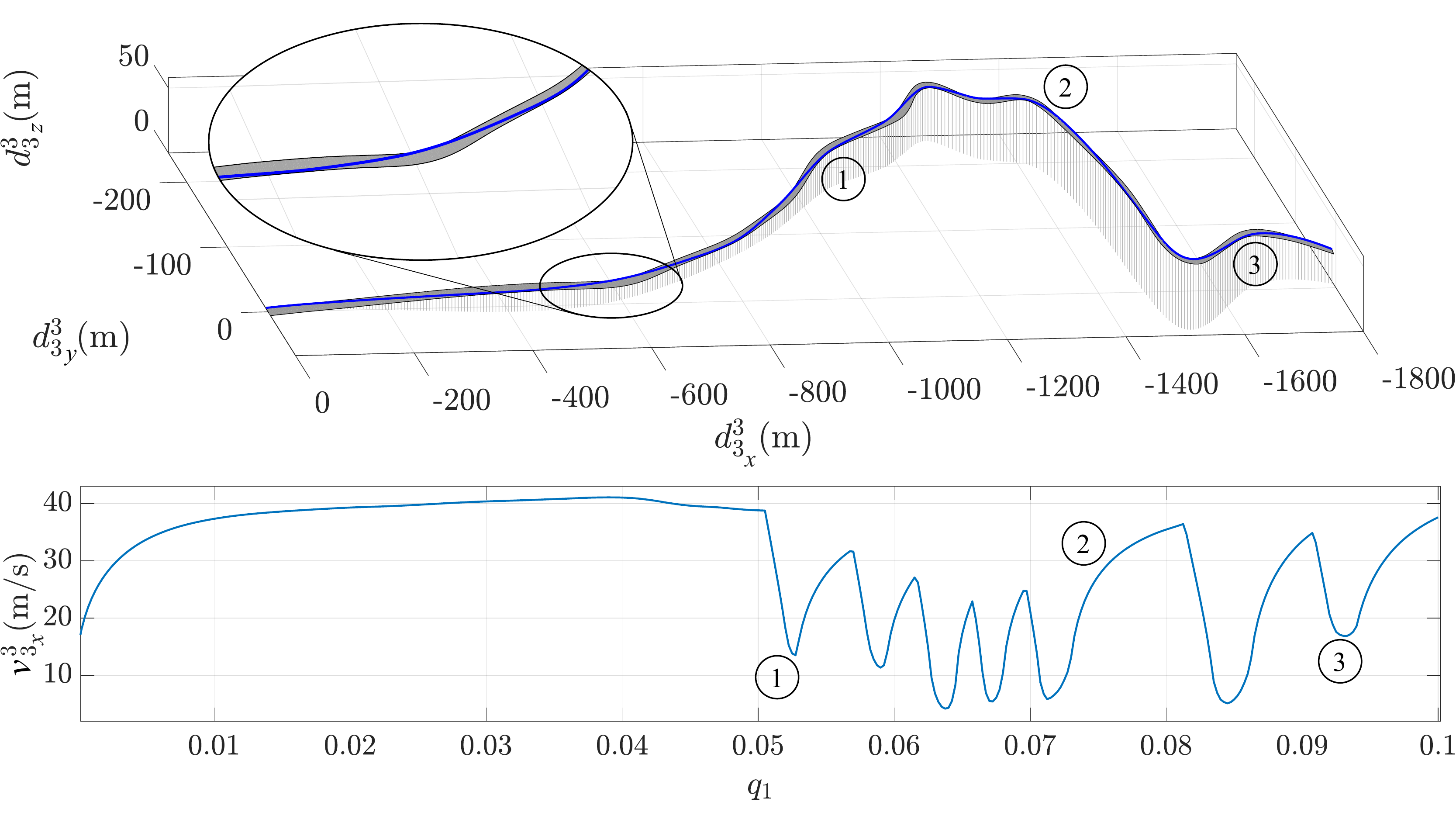}
	\caption{Optimal trajectory (top) and longitudinal velocity profile (bottom). As highlighted through the zoom, the optimizer cuts correctly the corners staying into the track limits.}
	\label{fig:traj}
\end{figure*}

Here, controls $u(q_1)$, states $x(q_1)$ and algebraic variables $z(q_1)$ are discretized on a fixed space grid $q_{1_i} = \Delta_q i$, ($i = 0,\dots, N$), with $\Delta_q = q_{1_N}/N$, where, $q_{1_N}$ is the final value of the spline parameter and $N$ is the number of mesh intervals. Hence, in agreement with the dimension of controls, states and algebraic vectors, we have that $u(q_{1_i}) = u_i \in \mathbb{R}^{3}$, $x(q_{1_i}) = x_i \in \mathbb{R}^{12}$ and $z(q_{1_i}) = z_i \in \mathbb{R}^{7}$.
Then, $v_i$ is used to indicate the \emph{collocation states}~\cite{Bartali} within each $i$-th interval.

Equality constraints $g_i(\cdot)$ include the dynamic equations~\eqref{eq:ode}, and the path algebraic equations~\eqref{eq:fijzf} and~\eqref{eq:congr} involving the $z$ variables.

Inequality constraints $h_i(\cdot)$ involve all path constraints limiting states, controls, and algebraic parameters. Power limits, adherence constraints and bounds on the lateral displacement $q_2$, necessary to remain within track bounds, are included in this form.
The terminal constraints $r(\cdot)$ can be considered in the case of a closed lap optimization for continuity purpose.
Finally, the cost function is approximated in each interval by a quadrature formula.
A typical stage cost $l_i$ is of the form
\begin{equation}
\label{eq:cost}
l_i = (\Delta_q/\dot{q}_{1_i} )^2 + K_\delta(\delta_{i+1} - \delta_{i}) + K_f(f_{xa_i}f_{xb_i}),
\end{equation}
where the first term penalizes lap time and the second penalizes abrupt variations of the steer angle through the weight $K_\delta$. Instead, the last term is introduced, with its weight $K_f$, as a relaxation for the \emph{complementary constraint} $f_{xa}f_{xb} = 0$. This constraint prevents traction and braking forces from acting simultaneously.

\subsection{Numerical Results}

The optimal control problem is coded in a scripting environment using
the MATLAB interface to the open-source CasADi framework~\cite{Andersson2019}, which provides
building blocks to efficiently formulate and solve large-scale optimization problems.

The optimal solution of the MLTP is obtained and discussed for a formula SAE vehicle (whose data are shown in Table~\ref{vehicle-data}) in two cases: i) for the first two kilometers of the \emph{N\"urburgring circuit} (which is, overall, $\simeq 21$ km long) and ii) for the full lap. 

\begin{table}
	\begin{center}
			\caption{The table lists the principal vehicle parameters and the corresponding numerical values (Par. and Value columns in the table), used to define the formula SAE car employed in the MLTP application. For the notation we refer to~\cite{Guiggiani}.}\label{vehicle-data}%
			\begin{tabular}{l|llll}
				\toprule
				 &Par. & Value & Par. & Value\\
				\midrule
\multirow{2}{4em}{Inertia} & $m$ & 240 kg & $I_{xx}$ & 40 kg/m$^2$\\
				& $I_{yy}$ & 100 kg/m$^2$ & $I_{zz}$ & 110 kg/m$^2$ \\
				\cmidrule(rl){1-5}
\multirow{4}{5em}{Geometric properties} & $a_1$ & 0.765 m  & $a_2$  & 0.815 m \\
				& $t_1$ & 1.21 m & $t_2$ & 1.21 m \\
				& $q_1$ & 0.335 m & $q_2$ & 0.335 m \\
				& $h$ & 0.435 m & & \\
				\cmidrule(rl){1-5}
\multirow{3}{5em}{Aerodynamic properties} & $S$ & 1.4 m$^2$ &  $C_x$ & 0.84 \\
				& $C_{z1}$  & 0.536 & $C_{z2}$ & 0.804 \\
				& $C_{z2}$ & 0.804 & &  \\	
				\cmidrule(rl){1-5} 	
\multirow{2}{5em}{Suspensions} & $K_{1i}$ & 36 kN/m & $C_{1i}$ & 3.28 kN/m/s \\	
				& $K_{2i}$ & 24 kN/m & $C_{2i}$ & 2.20 kN/m/s \\
				\cmidrule(rl){1-5}
\multirow{1}{5em}{PowerTrain}	& $P_{\text{max}}$ & 47 kW	& & \\
				\botrule
			\end{tabular}
	\end{center}
\end{table}

All the calculation times shown in the next sections, refer to a laptop
with 2.30 GHz Intel(R) Core(TM) i7-10875H CPU and 32 GB di RAM.

\subsubsection{N\"urburgring MLTP: First two kilometers}
To correctly capture vehicle dynamics, the number of discretization intervals is chosen equal to $N = 400$, leading to a total number of optimization variables $N_{\text{opt}} = 18412$ and to an optimal sector time $t_{\text{opt}} = 90.3$ s. Remarkably, the solution time for this problem was obatined in $t_c \simeq 38$ s with 64 iterations.

\begin{figure}
	\centering
	\includegraphics[width=1\linewidth]{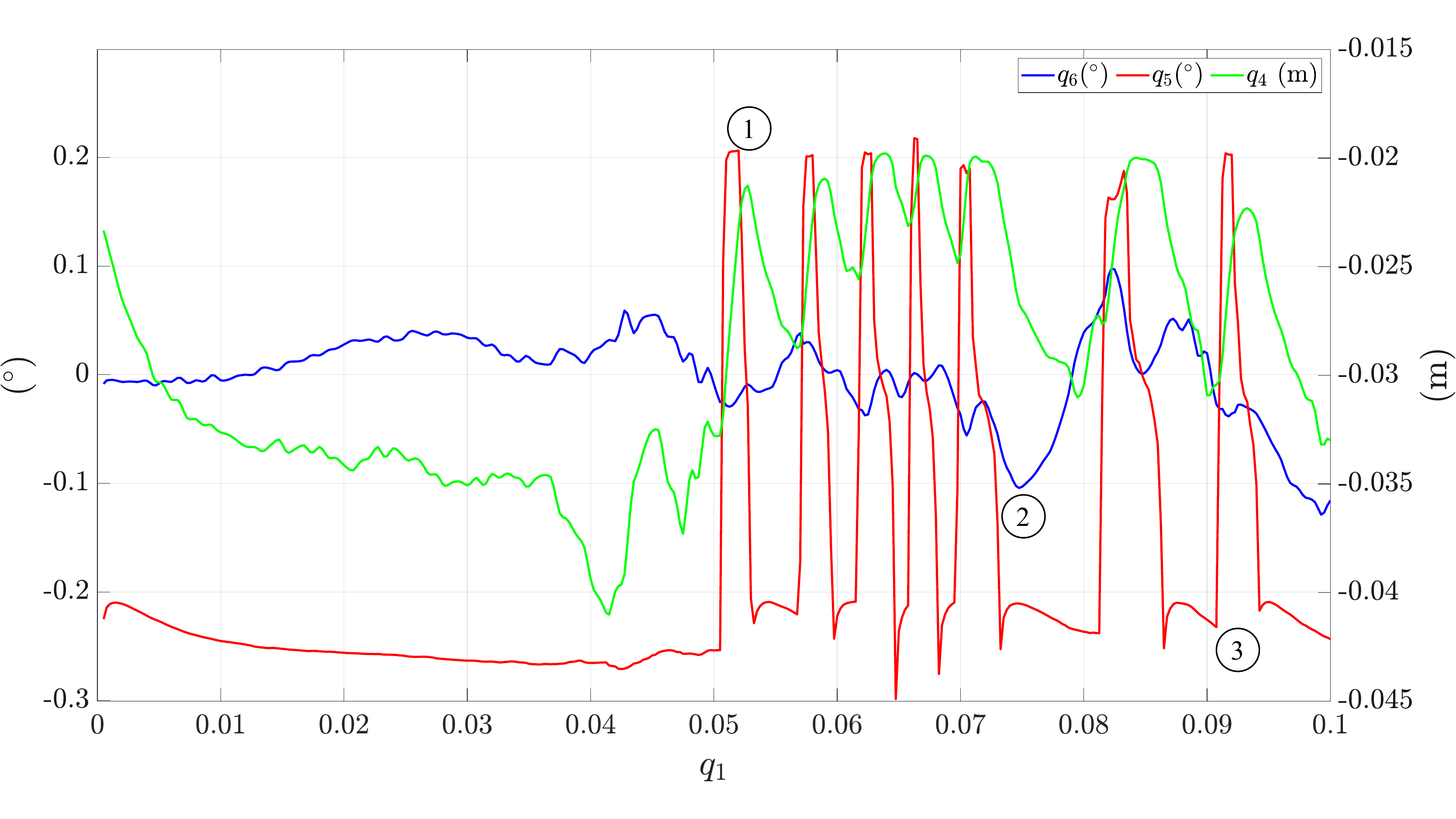}
	\caption{Optimal roll angle ($q_6$), pitch angle ($q_5$) and vertical suspension displacement ($q_4$). Considering the numbered portions of the track and the signals, it is evident that these signals are consistent with vehicle dynamics. As an example, the pitch angle $q_5$ increases and becomes positive when the vehicle is braking, see points 1 and 3; it is also worth observing that the roll angle $q_6$ remains negative for the duration of the high-speed curve 2 and changes sign before point 3.}
	\label{fig:angle}
\end{figure}

\begin{figure}
	\centering
	\includegraphics[width=1\linewidth]{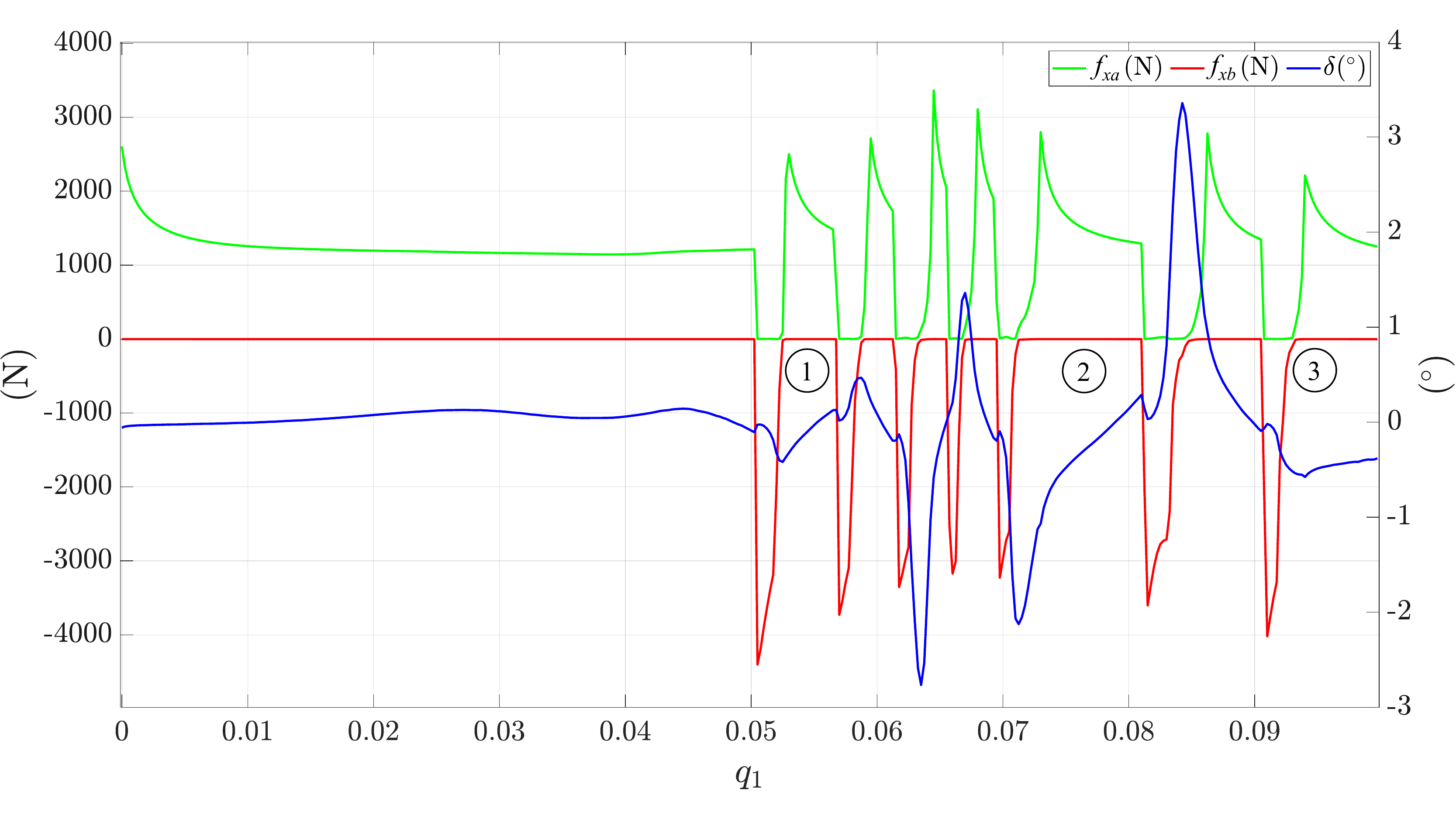}
	\caption{Optimal traction force ($f_{xa}$), braking force ($f_{xb}$) and wheel steer angle ($\delta$). As shown in this figure, the complementary constraint $f_{xa}f_{xb} = 0$ is correctly fulfilled, thus avoiding traction and breaking forces to act simultaneously.}
	\label{fig:controls}
\end{figure}

In Fig.~\ref{fig:traj} the optimal trajectory (starting at point $[0\,0\,0]^T$) and the optimal longitudinal speed profile $v^3_{3_x}$ are shown. Three different track portions are highlighted by numbered circles, in order to investigate the correct behaviour of the vehicle. In fact, as shown in Fig.~\ref{fig:angle}, the pitch angle $q_5$ increases and becomes positive when the vehicle is braking (see points 1 and 3), and decreases becoming negative when the vehicle accelerates (see point 2). A coherent behaviour also for the roll angle $q_6$ can be checked considering, for example, the last turn (point 3). Here, a negative steer angle shown in Fig.~\ref{fig:controls}  indicates a right turn and is consistent with a negative roll angle, as expected from vehicle dynamics.
Finally, Fig.~\ref{fig:controls} shows also that the complementarity constraint introduced in the cost function~\eqref{eq:cost} is fulfilled with sufficient accuracy.

\subsubsection{N\"urburgring MLTP: Full Lap}
 The optimal solution was also obtained for one lap of the whole \emph{N\"urburgring circuit} with noteworthy results. In this case, the discretization intervals are chosen equal to $N=1500$, which leads to a total number of optimization variables $N_{\text{opt}} = 69011$ and to an optimal lap time $t_{\text{opt}} = 666.3$ s. Remarkably, the solution time registered for this problem is $t_c \simeq 98$ s with only 49 iterations.

\begin{figure*}
	\centering
	\includegraphics[width=1\linewidth]{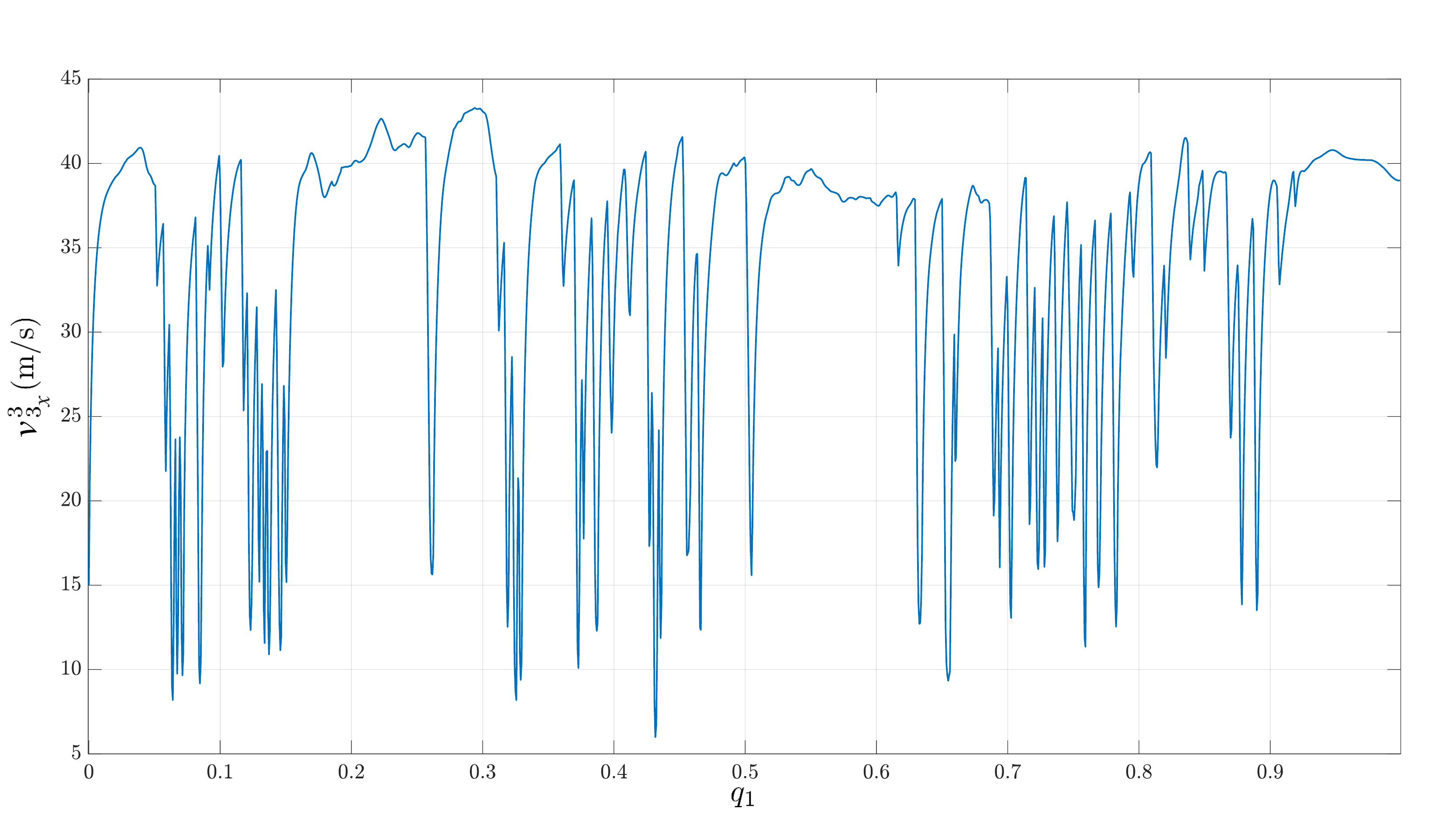}
	\caption{Optimal speed profile for the entire N\"urburgring lap.}
	\label{fig:u_lap}
\end{figure*}

\begin{figure*}
	\centering
	\includegraphics[width=1\linewidth]{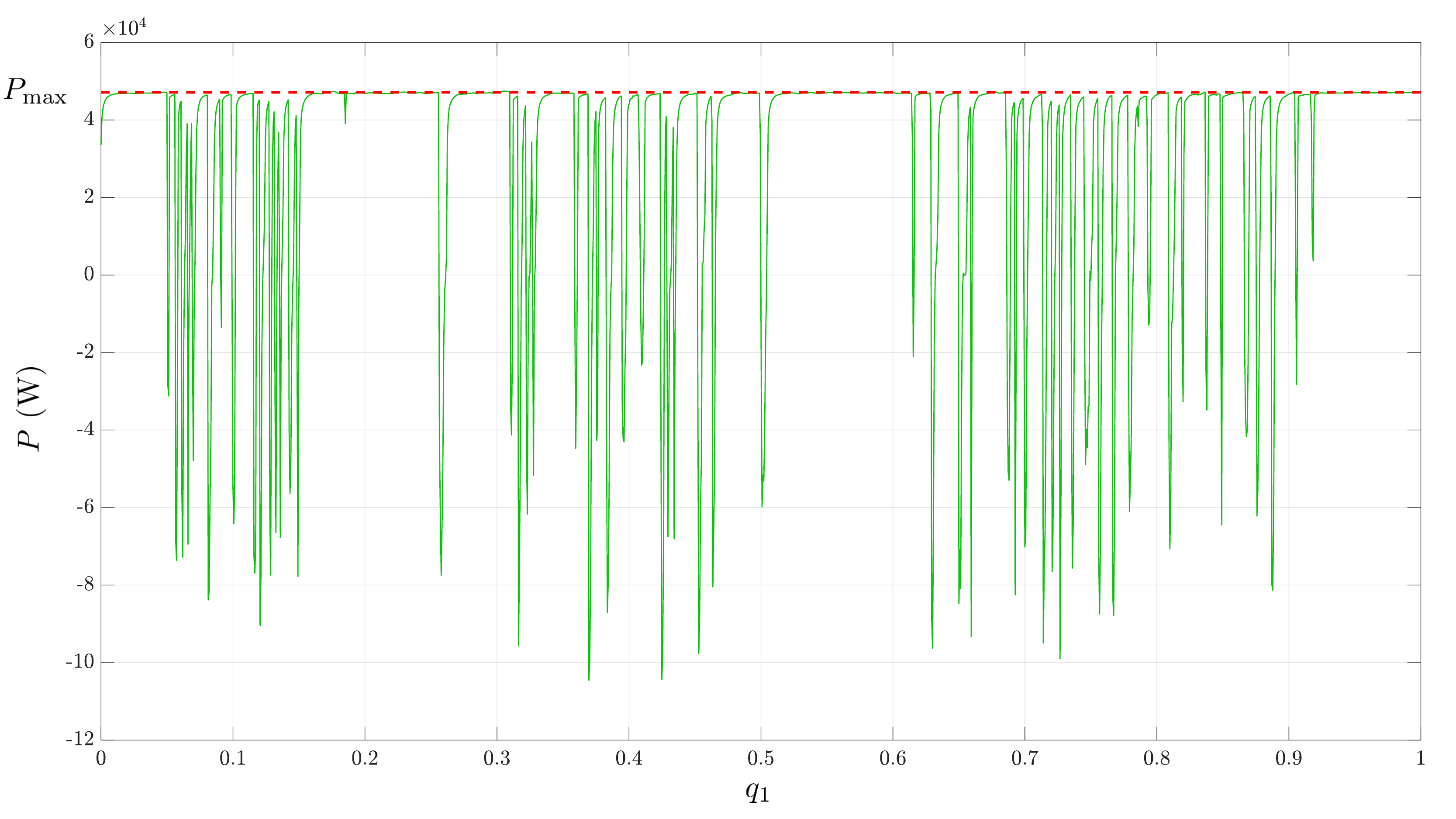}
	\caption{Optimal power profile for the entire N\"urburgring lap. The optimizer strives to stay as close as possible to the power limit minimizing lap time.}
	\label{fig:power_lap}
\end{figure*}

In Fig.~\ref{fig:u_lap} the optimal longitudinal speed profile is shown.
The plot density highlights the huge length of this track. Instead, in  Fig.~\ref{fig:power_lap}, the optimal power profile is shown. The red dashed line represents the limit on traction power due to the vehicle engine. As we can see the optimizer tries to stay as close as possible to this limit, in order to minimize the lap time.

\section{Conclusions}\label{sec:conclusions}
In this paper a Lie group-based race car model is presented. The vehicle is devised as a serial kinematic chain, linked to a 3D track with properly defined joints.

It is clearly exhibited that our framework merges gracefully with the Articulated Body Algorithm (ABA) and enables a fresh and systematic formulation of vehicle dynamics. A noteworthy contribution is represented by a rigorous reconciliation of the ABA steps with the salient features of vehicle dynamics, such as road-tire interactions, nonlinear tire characteristics, aerodynamic forces, and longitudinal and lateral load transfers. In particular, we discuss how the latter bring about the need to introduce algebraic variables and encode the dynamics as a DAE system.


To foster the validity of the presented approach, its application to set up the vehicle dynamics equations in an optimal planning problem is presented. In particular, two MLTPs, one for the first two kilometers and one for the whole lap of the N\"urburgring circuit, are setup within our framework. The obtained equations are successively discretized with the \emph{direct collocation} method and the solution is found within the CasADi suite. The results show that with the equations produced with our framework it is possible to very efficiently obtain correct optimal states and control inputs considering also bounce, roll and pitch motions. As a last remark, it is worth to point out that our framework opens up the possibility to directly employ efficient and open-source rigid body dynamics libraries, like~\cite{RBDL}, also within the vehicle dynamics context.

\section*{Declarations}
\subsection*{Funding}
No funding was received for conducting this study.
\subsection*{Conflict of Interest}
The authors declare that they have no conflict of interest.



\begin{thebibliography}{24}
\ifx \bisbn   \undefined \def \bisbn  #1{ISBN #1}\fi
\ifx \binits  \undefined \def \binits#1{#1}\fi
\ifx \bauthor  \undefined \def \bauthor#1{#1}\fi
\ifx \batitle  \undefined \def \batitle#1{#1}\fi
\ifx \bjtitle  \undefined \def \bjtitle#1{#1}\fi
\ifx \bvolume  \undefined \def \bvolume#1{\textbf{#1}}\fi
\ifx \byear  \undefined \def \byear#1{#1}\fi
\ifx \bissue  \undefined \def \bissue#1{#1}\fi
\ifx \bfpage  \undefined \def \bfpage#1{#1}\fi
\ifx \blpage  \undefined \def \blpage #1{#1}\fi
\ifx \burl  \undefined \def \burl#1{\textsf{#1}}\fi
\ifx \doiurl  \undefined \def \doiurl#1{\url{https://doi.org/#1}}\fi
\ifx \betal  \undefined \def \betal{\textit{et al.}}\fi
\ifx \binstitute  \undefined \def \binstitute#1{#1}\fi
\ifx \binstitutionaled  \undefined \def \binstitutionaled#1{#1}\fi
\ifx \bctitle  \undefined \def \bctitle#1{#1}\fi
\ifx \beditor  \undefined \def \beditor#1{#1}\fi
\ifx \bpublisher  \undefined \def \bpublisher#1{#1}\fi
\ifx \bbtitle  \undefined \def \bbtitle#1{#1}\fi
\ifx \bedition  \undefined \def \bedition#1{#1}\fi
\ifx \bseriesno  \undefined \def \bseriesno#1{#1}\fi
\ifx \blocation  \undefined \def \blocation#1{#1}\fi
\ifx \bsertitle  \undefined \def \bsertitle#1{#1}\fi
\ifx \bsnm \undefined \def \bsnm#1{#1}\fi
\ifx \bsuffix \undefined \def \bsuffix#1{#1}\fi
\ifx \bparticle \undefined \def \bparticle#1{#1}\fi
\ifx \barticle \undefined \def \barticle#1{#1}\fi
\bibcommenthead
\ifx \bconfdate \undefined \def \bconfdate #1{#1}\fi
\ifx \botherref \undefined \def \botherref #1{#1}\fi
\ifx \url \undefined \def \url#1{\textsf{#1}}\fi
\ifx \bchapter \undefined \def \bchapter#1{#1}\fi
\ifx \bbook \undefined \def \bbook#1{#1}\fi
\ifx \bcomment \undefined \def \bcomment#1{#1}\fi
\ifx \oauthor \undefined \def \oauthor#1{#1}\fi
\ifx \citeauthoryear \undefined \def \citeauthoryear#1{#1}\fi
\ifx \endbibitem  \undefined \def \endbibitem {}\fi
\ifx \bconflocation  \undefined \def \bconflocation#1{#1}\fi
\ifx \arxivurl  \undefined \def \arxivurl#1{\textsf{#1}}\fi
\csname PreBibitemsHook\endcsname

\bibitem{Massaro-general}
\begin{barticle}
\bauthor{\bsnm{Massaro}, \binits{M.}},
\bauthor{\bsnm{Limebeer}, \binits{D.J.N.}}:
\batitle{Minimum-lap-time optimisation and simulation}.
\bjtitle{Vehicle System Dynamics}
\bvolume{59}(\bissue{7}),
\bfpage{1069}--\blpage{1113}
(\byear{2021})
\end{barticle}
\endbibitem

\bibitem{Track-Ribbon}
\begin{botherref}
\oauthor{\bsnm{Perantoni}, \binits{G.}},
\oauthor{\bsnm{Limebeer}, \binits{D.J.N.}}:
Optimal {C}ontrol of a {F}ormula {O}ne {C}ar on a {T}hree-{D}imensional {T}rack
  {P}art 1: {T}rack {M}odeling and {I}dentification.
Journal of Dynamic Systems, Measurement, and Control
\textbf{137}(5)
(2015)
\end{botherref}
\endbibitem

\bibitem{Track-Ribbon-camber}
\begin{botherref}
\oauthor{\bsnm{Lovato}, \binits{S.}},
\oauthor{\bsnm{Massaro}, \binits{M.}},
\oauthor{\bsnm{Limebeer}}:
Curved-ribbon-based track modelling for minimum lap-time optimisation.
Meccanica
(56)
(2021)
\end{botherref}
\endbibitem

\bibitem{Guiggiani}
\begin{bbook}
\bauthor{\bsnm{Guiggiani}, \binits{M.}}:
\bbtitle{The {S}cience of {V}ehicle {D}ynamics: {H}andling, {B}raking, and
  {R}ide of {R}oad and {R}ace {C}ars},
\bedition{3rd} edn.
\bpublisher{Springer},
\blocation{Cham}
(\byear{2018})
\end{bbook}
\endbibitem

\bibitem{SingleTrack}
\begin{bchapter}
\bauthor{\bsnm{Rucco}, \binits{A.}},
\bauthor{\bsnm{Notarstefano}, \binits{G.}},
\bauthor{\bsnm{Hauser}, \binits{J.}}:
\bctitle{Computing minimum lap-time trajectories for a single-track car with
  load transfer}.
In: \bbtitle{51st IEEE Conference on Decision and Control (CDC)},
pp. \bfpage{6321}--\blpage{6326}
(\byear{2012})
\end{bchapter}
\endbibitem

\bibitem{Double-friction}
\begin{barticle}
\bauthor{\bsnm{Christ}, \binits{F.}},
\bauthor{\bsnm{Wischnewski}, \binits{A.}},
\bauthor{\bsnm{Heilmeier}, \binits{A.}},
\bauthor{\bsnm{Lohmann}, \binits{B.}}:
\batitle{Time-optimal trajectory planning for a race car considering variable
  tyre-road friction coefficients}.
\bjtitle{Vehicle System Dynamics}
\bvolume{59}(\bissue{4}),
\bfpage{588}--\blpage{612}
(\byear{2021})
\end{barticle}
\endbibitem

\bibitem{Pac}
\begin{bbook}
\bauthor{\bsnm{Pacejka}, \binits{H.}}:
\bbtitle{Tire and Vehicle Dynamics},
\bedition{2nd} edn.
\bpublisher{Butterworth-Heinemann},
\blocation{London}
(\byear{2002})
\end{bbook}
\endbibitem

\bibitem{Double-aero}
\begin{barticle}
\bauthor{\bparticle{de} \bsnm{Buck}, \binits{P.}},
\bauthor{\bsnm{Martins}, \binits{J.R.R.A.}}:
\batitle{Minimum lap time trajectory optimisation of performance vehicles with
  four-wheel drive and active aerodynamic control}.
\bjtitle{Vehicle System Dynamics}
\bvolume{0}(\bissue{0}),
\bfpage{1}--\blpage{17}
(\byear{2022})
\end{barticle}
\endbibitem

\bibitem{Double-LSD}
\begin{botherref}
\oauthor{\bparticle{van} \bsnm{Koutrik}, \binits{S.}}:
Optimal control for race car minimum time maneuvering.
Master thesis,
Delft University of Technology,
Faculty of Mechanical, Maritime and Materials Engineering (3mE)
(2015)
\end{botherref}
\endbibitem

\bibitem{Double-F1}
\begin{botherref}
\oauthor{\bsnm{Limebeer}, \binits{D.J.N.}},
\oauthor{\bsnm{Perantoni}, \binits{G.}}:
Optimal control of a formula one car on a three-dimensional track part 2:
  Optimal control.
Journal of Dynamic Systems, Measurement, and Control
\textbf{137}(5)
(2015)
\end{botherref}
\endbibitem

\bibitem{Multibody-planar}
\begin{botherref}
\oauthor{\bsnm{Ambr{\'o}sio}, \binits{J.}},
\oauthor{\bsnm{Marques}, \binits{L.}}:
Optimal lap time for a race car: A planar multibody dynamics approach.
Interdisciplinary Applications of Kinematics
(2019)
\end{botherref}
\endbibitem

\bibitem{Multibody-3D}
\begin{barticle}
\bauthor{\bsnm{Bianco}, \binits{N.D.}},
\bauthor{\bsnm{Lot}, \binits{R.}},
\bauthor{\bsnm{Gadola}, \binits{M.}}:
\batitle{Minimum time optimal control simulation of a {GP2} race car}.
\bjtitle{Proceedings of the Institution of Mechanical Engineers, Part D:
  Journal of Automobile Engineering}
\bvolume{232},
\bfpage{1180}--\blpage{1195}
(\byear{2018})
\end{barticle}
\endbibitem

\bibitem{Mueller2003}
\begin{barticle}
\bauthor{\bsnm{Mueller}, \binits{A.}},
\bauthor{\bsnm{Maisser}, \binits{P.}}:
\batitle{A {Lie-Group} formulation of kinematics and dynamics of constrained
  mbs and its application to analytical mechanics}.
\bjtitle{Multibody System Dynamics}
\bvolume{9}(\bissue{4}),
\bfpage{311}--\blpage{352}
(\byear{2003})
\end{barticle}
\endbibitem

\bibitem{Featherstone}
\begin{bbook}
\bauthor{\bsnm{Featherstone}, \binits{R.}}:
\bbtitle{Rigid Body Dynamics Algorithms}.
\bpublisher{Springer},
\blocation{Heidelberg}
(\byear{2007})
\end{bbook}
\endbibitem

\bibitem{RBDL}
\begin{botherref}
\oauthor{\bsnm{Felis}, \binits{M.L.}}:
{RBDL}: Rigid Body Dynamics Library.
\url{https://github.com/rbdl/rbdl}
(2023)
\end{botherref}
\endbibitem

\bibitem{Felis2016}
\begin{botherref}
\oauthor{\bsnm{Felis}, \binits{M.L.}}:
Rbdl: an efficient rigid-body dynamics library using recursive algorithms.
Autonomous Robots,
1--17
(2016).
\doiurl{10.1007/s10514-016-9574-0}
\end{botherref}
\endbibitem

\bibitem{Muller}
\begin{barticle}
\bauthor{\bsnm{M{\"u}ller}, \binits{A.}}:
\batitle{Higher derivatives of the kinematic mapping and some applications}.
\bjtitle{Mechanism and Machine Theory}
\bvolume{76},
\bfpage{70}--\blpage{85}
(\byear{2014})
\end{barticle}
\endbibitem

\bibitem{Sastry}
\begin{bbook}
\bauthor{\bsnm{Murray}, \binits{R.M.}},
\bauthor{\bsnm{Li}, \binits{Z.}},
\bauthor{\bsnm{Sastry}, \binits{S.S.}}:
\bbtitle{A Mathematical Introduction to Robotic Manipulation}.
\bpublisher{CRC Press},
\blocation{Boca Raton}
(\byear{1994})
\end{bbook}
\endbibitem

\bibitem{nurbs}
\begin{bbook}
\bauthor{\bsnm{Piegl}, \binits{L.}},
\bauthor{\bsnm{Tiller}, \binits{W.}}:
\bbtitle{The NURBS Book}.
\bpublisher{Springer},
\blocation{Berlin}
(\byear{1995})
\end{bbook}
\endbibitem

\bibitem{Lot-time}
\begin{bchapter}
\bauthor{\bsnm{Lot}, \binits{R.}},
\bauthor{\bsnm{Bianco}, \binits{N.D.}}:
\bctitle{The significance of high-order dynamics in lap time simulations}.
In: \bbtitle{IAVSD 2015: 24th International Symposium on Dynamics of Vehicles
  on Roads and Tracks (16/08/15 - 20/08/15)}
(\byear{2015})
\end{bchapter}
\endbibitem

\bibitem{Lot-spatial}
\begin{barticle}
\bauthor{\bsnm{Lot}, \binits{R.}},
\bauthor{\bsnm{Biral}, \binits{F.}}:
\batitle{A curvilinear abscissa approach for the lap time optimization of
  racing vehicles}.
\bjtitle{IFAC Proceedings Volumes}
\bvolume{47}(\bissue{3}),
\bfpage{7559}--\blpage{7565}
(\byear{2014})
\end{barticle}
\endbibitem

\bibitem{Betts}
\begin{bbook}
\bauthor{\bsnm{Betts}, \binits{J.T.}}:
\bbtitle{Practical Methods for Optimal Control and Estimation Using Nonlinear
  Programming, Second Edition},
\bedition{3rd} edn.
\bpublisher{SIAM - Society for Industrial and Applied Mathematics},
\blocation{Philadelphia}
(\byear{2010})
\end{bbook}
\endbibitem

\bibitem{Bartali}
\begin{barticle}
\bauthor{\bsnm{Gabiccini}, \binits{M.}},
\bauthor{\bsnm{Bartali}, \binits{L.}},
\bauthor{\bsnm{Guiggiani}, \binits{M.}}:
\batitle{Analysis of {D}riving {S}tyles of a {GP2} {C}ar {v}ia {M}inimum
  {L}ap-{T}ime {D}irect {T}rajectory {O}ptimization}.
\bjtitle{Multibody System Dynamics}
\bvolume{53},
\bfpage{85}--\blpage{113}
(\byear{2021})
\end{barticle}
\endbibitem

\bibitem{Andersson2019}
\begin{barticle}
\bauthor{\bsnm{Andersson}, \binits{J.A.E.}},
\bauthor{\bsnm{Gillis}, \binits{J.}},
\bauthor{\bsnm{Horn}, \binits{G.}},
\bauthor{\bsnm{Rawlings}, \binits{J.B.}},
\bauthor{\bsnm{Diehl}, \binits{M.}}:
\batitle{{CasADi} -- {A} software framework for nonlinear optimization and
  optimal control}.
\bjtitle{Mathematical Programming Computation}
\bvolume{11}(\bissue{1}),
\bfpage{1}--\blpage{36}
(\byear{2019}).
\doiurl{10.1007/s12532-018-0139-4}
\end{barticle}
\endbibitem

\end{thebibliography}


\end{document}